\documentclass{article}

\usepackage{microtype}
\usepackage{graphicx}
\usepackage{subcaption}
\usepackage{amsmath}     
\usepackage{booktabs}
\usepackage{siunitx}
\usepackage{longtable}
\usepackage[table]{xcolor}
\usepackage{ragged2e} 

\usepackage{amsfonts, amssymb}
\usepackage{algorithm2e}
\usepackage[most]{tcolorbox}
\usepackage{tabularx}
\usepackage{hyperref}

\usepackage[preprint]{icml2026} 

\usepackage{amsmath}
\usepackage{amssymb}
\usepackage{mathtools}
\usepackage{amsthm}
\definecolor{boxBack}{RGB}{245,245,245}
\definecolor{algBlue}{RGB}{30, 90, 160}
\definecolor{algOrange}{RGB}{230, 120, 20}
\usepackage[capitalize,noabbrev]{cleveref}

\theoremstyle{plain}

\theoremstyle{definition}

\theoremstyle{remark}

\usepackage[textsize=tiny]{todonotes}

\icmltitlerunning{Preprint}

\begin{document}

\twocolumn[
  \icmltitle{Barycentric alignment for instance-level comparison of neural representations}


  \icmlsetsymbol{equal}{*}

  \begin{icmlauthorlist}
    \icmlauthor{Shreya Saha}{equal,yyy}
    \icmlauthor{Zoe Wanying He}{equal,comp}
    \icmlauthor{Meenakshi Khosla}{comp,sch}
  \end{icmlauthorlist}

  \icmlaffiliation{yyy}{Department of Electrical and Computer Engineering, UCSD}
  \icmlaffiliation{comp}{Cognitive Science Department, UCSD}
  \icmlaffiliation{sch}{Department of Computer Science and Engineering, UCSD}

  \icmlcorrespondingauthor{Shreya Saha}{ssaha@ucsd.edu}
  \icmlcorrespondingauthor{Zoe He}{wah016@ucsd.edu}

  \icmlkeywords{Machine Learning, ICML}

  \vskip 0.3in
]



\printAffiliationsAndNotice{\icmlEqualContribution}

\begin{figure}[h!]
\centering
\includegraphics[width=0.8\linewidth, trim=3cm 0cm 5cm 0cm, clip]{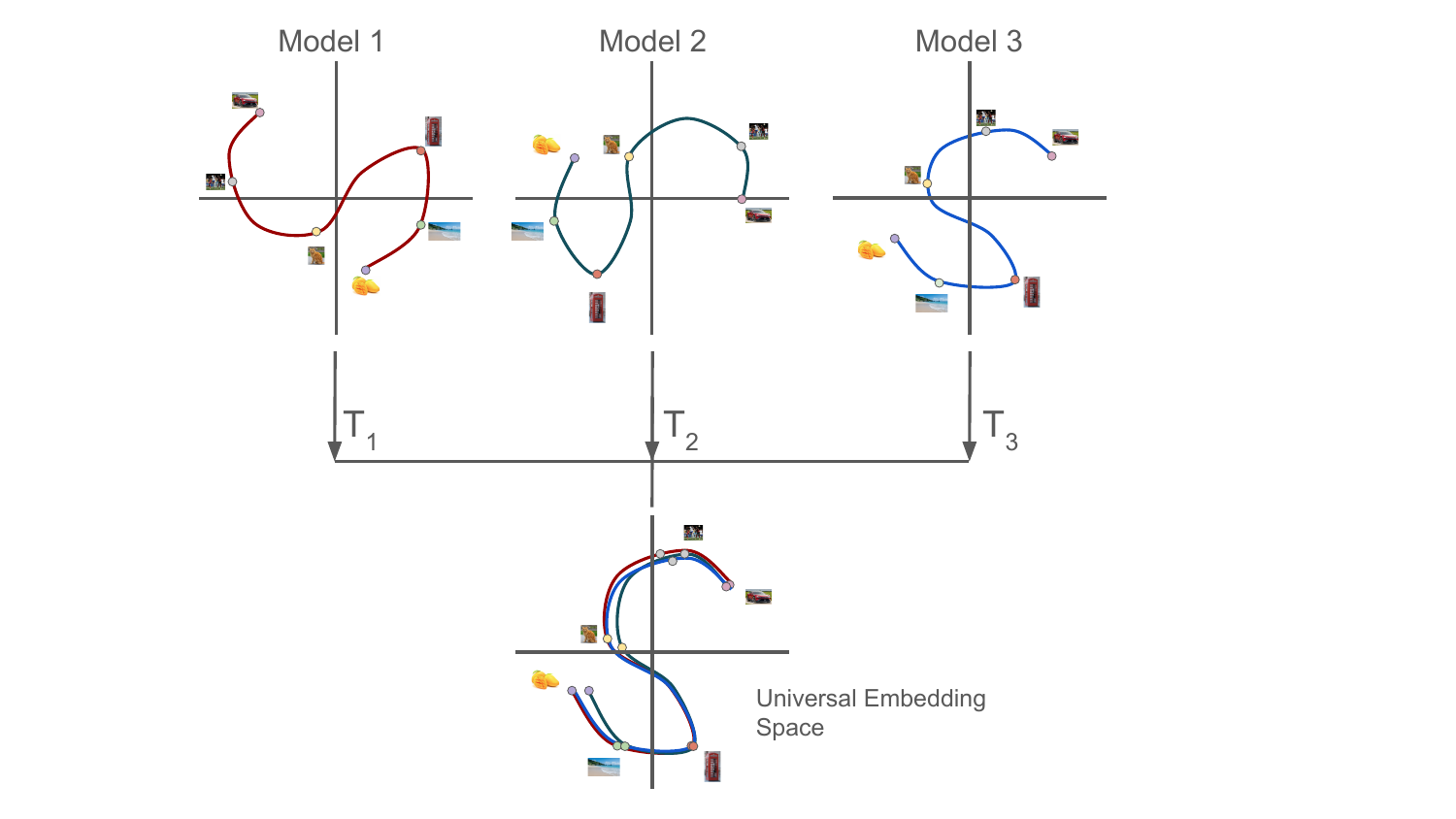}
\caption{\textbf{Barycentric alignment into a universal embedding space.} Representations from a diverse set of models, spanning architectures, training objectives, and scales, are aligned via a barycentric procedure into a shared embedding space by factoring out nuisance symmetries. }
\label{fig:fig3}
\end{figure}

\begin{abstract}
Comparing representations across neural networks is challenging because representations admit symmetries, such as arbitrary reordering of units or rotations of activation space, that obscure underlying equivalence between models. We introduce a barycentric alignment framework that quotients out these nuisance symmetries to construct a universal embedding space across many models. Unlike existing similarity measures, which summarize relationships over entire stimulus sets, this framework enables similarity to be defined at the level of individual stimuli, revealing inputs that elicit convergent versus divergent representations across models. Using this instance-level notion of similarity, we identify systematic input properties that predict representational convergence versus divergence across vision and language model families. We also construct universal embedding spaces for brain representations across individuals and cortical regions, enabling instance-level comparison of representational agreement across stages of the human visual hierarchy. Finally, we apply the same barycentric alignment framework to purely unimodal vision and language models and find that post-hoc alignment into a shared space yields image–text similarity scores that closely track human cross-modal judgments and approach the performance of contrastively trained vision-language models. This strikingly suggests that independently learned representations already share sufficient geometric structure for human-aligned cross-modal comparison. Together, these results show that resolving representational similarity at the level of individual stimuli reveals phenomena that cannot be detected by set-level comparison metrics.
  
\end{abstract}

\section{Introduction}

Across domains, deep neural networks map stimuli into high-dimensional internal representations that support perception, reasoning, and action. Despite their central role, our understanding of these representations remains limited. Most analyses probe individual models or narrow architectural families, identifying patterns or mechanisms that explain behavior in a specific system. This leaves open a fundamental scientific question: when a representational phenomenon is observed in a single network, which aspects reflect idiosyncratic implementation choices, and which reflect shared representational structure that persists across architectures, training regimes, scales, or even modalities? As models continue to diversify, the field lacks principled tools for determining where universality ends and model-specificity begins in neural representations.

Addressing this question requires tools for comparing representations across models, rather than explaining them in isolation. A substantial body of prior work has therefore focused on representational similarity analysis, introducing measures such as RSA \cite{rsa}, CKA \cite{kornblith2019similarity}, SVCCA \cite{raghu2017svcca}, and related metrics to quantify alignment between neural networks (\cite{yamins2014performance}, \cite{laakso2022can}, Li et al., 2015; \cite{morcos2018insights},  Wang et al., 2018). 
These approaches have yielded important insights into how models relate with one another, revealing broad trends across architectures, training regimes, and layers. However, they fundamentally characterize representational similarity as a set-level property: a single score that summarizes how an entire collection of stimuli is arranged in two representation spaces. In doing so, they obscure how agreement and disagreement are distributed across the input space.  
This limitation precludes a finer-grained view of representational structure - one that treats alignment as an instance-level phenomenon and examines its variation across stimuli and model families. Such a view is essential for characterizing the \emph{ecology of representational alignment}: which inputs induce consensus and which expose divergence. Moreover, because most existing approaches operate pairwise, they do not provide a principled way to align or analyze representations from an entire population of models simultaneously. As a result, we currently lack tools for defining stimulus-level notions of agreement that are stable, interpretable, and comparable across large model ensembles.

A small number of recent studies have begun to move beyond purely aggregate comparison, but important limitations remain (e.g., \cite{hosseini2024universality}, \cite{kollinginvestigating}). Hosseini et al. partition a fixed stimulus set into “high-agreement” and “low-agreement” subsets and show that stimuli on which artificial models disagree most strongly are also those on which models and human brain representations diverge. However, whether a particular stimulus is labeled “agreeable” depends on the composition of the evaluation set: the same item can appear highly consistent or highly divergent depending on which other stimuli are present. Agreement therefore reflects not only intrinsic properties of a stimulus, but also contextual contrasts induced by the stimulus pool, undermining a stimulus-specific interpretation. Pointwise Normalised Kernel Alignment (PNKA; \cite{kollinginvestigating}) takes an important step toward finer granularity by assigning scores to individual stimuli. Yet PNKA still defines similarity relative to the global neighborhood structure of the evaluation set and remains inherently pairwise, with scores tied to a specific pair of models. As a result, extending the analysis to larger model populations or to new stimuli requires recomputing kernels, limiting its utility for studying representational agreement at the level of model ensembles.

More broadly, these limitations mean that the field lacks a systematic account of how representational agreement is distributed across the space of inputs. As a result, the ecology of representational similarity, particularly how agreement and disagreement are structured across stimuli, how they depend on input properties, and how they vary across model populations, remains surprisingly underexplored.

Here we introduce a barycentric framework for representational alignment that treats symmetry-quotienting as a first-class design principle, with Procrustes barycenters providing a natural solution for rotational invariances. This alignment enables a stimulus-specific measure of representational agreement, defined by how tightly a stimulus’s representations cluster across models in the shared space. Unlike prior approaches, this instance-wise measure is independent of the surrounding stimulus set and well-defined for populations of models rather than model pairs. Using this framework, we characterize which inputs elicit convergent versus divergent representations across vision and language models, construct universal embedding spaces for brain representations across individuals and regions, and show that post-hoc alignment of unimodal vision and language models recovers meaningful image–text similarity. Together, these results demonstrate how instance-level analysis reveals structure in representational similarity that is inaccessible to set-level comparison metrics.


\section{Method}
We propose a \emph{barycentric alignment} framework to map representations from many models into a single shared embedding space while explicitly quotienting out \emph{nuisance symmetries}. Each model’s representation of the same stimulus set is treated as a point cloud whose coordinates are only identifiable up to transformations induced by architectural and training choices (e.g., rotations). By aligning all models to a common \emph{barycenter} under an appropriate invariance class, we obtain a universal space in which representations become directly comparable across models and can be compared at the level of individual stimuli.

\subsection{Symmetries and the choice of alignment}

We treat two representations as equivalent if they are related by a transformation from a specified group $\mathcal{G}$. Formally, let $X_i \in \mathbb{R}^{n \times d}$ denote the representation matrix from model $i$, where rows correspond to $n$ stimuli and columns to $d$ features. We consider $X_i \sim X_j$ if and only if there exists a transformation $T \in \mathcal{G}$ such that $X_i$ can be approximated well with $X_j T$.

The choice of $\mathcal{G}$ reflects assumptions about which transformations preserve meaningful representational content. Permutation invariance ($\mathcal{G} = \mathcal{P}(d)$, the group of permutation matrices) captures the intuition that neuron ordering is arbitrary. Orthogonal invariance ($\mathcal{G} = O(d)$, the orthogonal group) additionally accounts for rotations and reflections of the activation space. Throughout this work, we adopt orthogonal invariance as our default equivalence relation, as it provides a principled middle ground: it quotients out rotational degrees of freedom while preserving the metric structure of the representation space. Critically, the orthogonal group consists of linear isometries, i.e., transformations that preserve Euclidean distances. This ensures that aligning representations via orthogonal transformations does not distort the geometric relationships among stimuli within each model's space.

\subsection{Barycentric Alignment via Procrustes Iteration}

Given $N$ models, each producing representations for the same set of $n$ training stimuli, we seek a common embedding space and a set of alignment transformations $\{T_i\}_{i=1}^N$ such that the aligned representations $\{X_i T_i\}$ are mutually consistent. We formulate this as the problem of computing the \emph{Procrustes barycenter} of the $N$ representation matrices, i.e., the mean shape in the quotient space induced by orthogonal equivalence. 

To ensure consistent dimensionality across models, we set $d = \max(d_1, \ldots, d_N)$, where $d_i$ denotes the original dimensionality of model $i$. Representations from models with $d_i < d$ are zero-padded to match the maximum dimensionality. We initialize the template $M^{(0)}$ as the arithmetic mean of the representations and then iterate the following procedure:

\textbf{Alignment step:} For each model $i$, compute the orthogonal Procrustes transformation $ T_i = \arg\min_{R \in O(d)} \|X_i R - M^{(t)}\|_F$,
    which admits a closed-form solution via the singular value decomposition.
    
\textbf{Update step:} Set $X_i^{(t)} \leftarrow X_i T_i$ and recompute the template as $M^{(t+1)} = \frac{1}{N} \sum_{i=1}^N X_i^{(t)}$.

This alternating procedure converges to a local minimum of the sum of squared Procrustes distances, yielding aligned representations that reside in a shared coordinate system. We stop early when $\|M^{(t+1)}-M^{(t)}\|_F/\|M^{(t)}\|_F<\varepsilon$. The resulting template $M$ can be interpreted as the Fr\'{e}chet mean of the $N$ representations under the Procrustes metric, the resulting barycenter coordinates constitute what we refer to as the \emph{universal embedding space} and $\{T_i\}$ provide model-specific maps into that space.

Algorithm~\ref{alg:procrustes-barycenter-training} summarizes the training phase. The $n$ training stimuli used to learn the alignment transformations are held separate from the test stimuli used for inference.

\subsection{Inference and stimulus-wise agreement.}
Given a new set of $m$ test stimuli, each model produces representations $Y_i \in \mathbb{R}^{m \times d}, \qquad i=1,\dots,N$,
which we project into the universal space via the learned transforms: $Y_i' = Y_i T_i$.
For each test stimulus $j\in\{1,\dots,m\}$, we then quantify representational agreement by averaging pairwise similarity across the aligned representations: $S_j = \frac{1}{N(N-1)}\sum_{p\neq q} \mathrm{SIM}\!\left(Y'_{p j},\, Y'_{q j}\right)$,
where $Y'_{p j}$ denotes the row of $Y_p'$ corresponding to stimulus $j$. The similarity function is left generic; unless otherwise stated, we use cosine similarity throughout.
Throughout, we refer to this stimulus-wise representational agreement as the \emph{instance-level representational consistency score}, which quantifies the extent to which a stimulus admits a consistent representation across models after quotienting out nuisance symmetries.

\begin{table}[H]
\centering
\caption{Cross-model correlation, RMS score, and stimulus retrieval accuracy for the supervised vision model pool and the language model pool.}
\label{tab:table1}
\small
\setlength{\tabcolsep}{4pt}

\begin{tabular}{@{}l 
                S[table-format=1.4] 
                S[table-format=1.5] 
                S[table-format=1.4] 
                S[table-format=1.4] 
                S[table-format=1.4]@{}}
\toprule
\textbf{Model} & {\textbf{Corr. $\uparrow$}} & {\textbf{RMS $\downarrow$}} & {\textbf{Top-1 $\uparrow$}} & {\textbf{Top-5 $\uparrow$}} & {\textbf{Top-10 $\uparrow$}} \\
\midrule

\multicolumn{6}{c}{\textit{Supervised Vision Model Pool}} \\
\midrule

ViT-B/16  & 0.4396 & 0.01344 & 0.7726 & 0.92493 & 0.9436 \\
ViT-S/16  & 0.4469 & 0.01333 & 0.7960 & 0.9259  & 0.9422 \\
ViT-T/16  & 0.3820 & 0.01422 & 0.6440 & 0.8242  & 0.8679 \\
ViT-L/16  & 0.4451 & 0.01336 & 0.6833 & 0.8956  & 0.9275 \\

ConvNeXt T & 0.4189 & 0.01349 & 0.7328 & 0.9063 & 0.9306 \\
ConvNeXt S & 0.4697 & 0.01281 & 0.7179 & 0.9061 & 0.9315 \\
ConvNeXt B & 0.4731 & 0.01269 & 0.6944 & 0.8977 & 0.9288 \\
ConvNeXt L & 0.4698 & 0.01285 & 0.6491 & 0.8826 & 0.9225 \\

Swin-T & 0.4891 & 0.01274 & 0.8143 & 0.9330 & 0.9459 \\
Swin-B & 0.4870 & 0.01271 & 0.6870 & 0.9002 & 0.9338 \\
Swin-S & 0.4807 & 0.01281 & 0.7132 & 0.9101 & 0.9382 \\

ResNet-18  & 0.4859 & 0.01287 & 0.8304 & 0.9344 & 0.9398 \\
ResNet-34  & 0.4976 & 0.01268 & 0.8470 & 0.9401 & 0.9449 \\
ResNet-50  & 0.5270 & 0.01225 & 0.8672 & 0.9251 & 0.9477 \\
ResNet-101 & 0.5282 & 0.01223 & 0.8636 & 0.9401 & 0.9477 \\
ResNet-152 & 0.5292 & 0.01222 & 0.8581 & 0.9399 & 0.9476 \\

VGG-11 & 0.4484 & 0.01375 & 0.8264 & 0.9305 & 0.9436 \\
VGG-13 & 0.4452 & 0.01381 & 0.8227 & 0.9318 & 0.9444 \\
VGG-16 & 0.4561 & 0.01362 & 0.8116 & 0.9308 & 0.9445 \\
VGG-19 & 0.4601 & 0.01354 & 0.7993 & 0.9291 & 0.9440 \\

AlexNet & 0.3495 & 0.01535 & 0.7576 & 0.8952 & 0.9209 \\

\midrule
\multicolumn{6}{c}{\textit{Autoregressive Language Model Pool}} \\
\midrule

OpenLLaMA-3b   & 0.5231 & 0.01094 & 0.9491 & 0.9679 & 0.9761 \\
OpenLLaMA-7b   & 0.4942 & 0.01162 & 0.8988 & 0.9309 & 0.9385 \\
OpenLLaMA-13b  & 0.4795 & 0.01171 & 0.9406 & 0.9721 & 0.9785 \\

Qwen2.5-0.5b  & 0.5260 & 0.01129 & 0.9261 & 0.9721 & 0.9815 \\
Qwen2.5-1.5b  & 0.5257 & 0.00985 & 0.5479 & 0.6573 & 0.7136 \\
Qwen2.5-3b    & 0.5323 & 0.01115 & 0.9215 & 0.9648 & 0.9742 \\
Qwen2.5-7b    & 0.4929 & 0.01015 & 0.6518 & 0.7406 & 0.7836 \\
Qwen2.5-14b   & 0.4771 & 0.01044 & 0.6991 & 0.7727 & 0.8161 \\

Llama3-1b  & 0.5781 & 0.00958 & 0.8570 & 0.9209 & 0.9448 \\
Llama3-3b  & 0.5483 & 0.01030 & 0.9491 & 0.9736 & 0.9833 \\
Llama3-8b  & 0.5170 & 0.01098 & 0.9727 & 0.9864 & 0.9927 \\

Gemma2-2b  & 0.5450 & 0.01098 & 0.9670 & 0.9861 & 0.9885 \\

\bottomrule
\end{tabular}
\end{table}

We evaluate the quality of the learned universal embedding by measuring cross-model representational alignment after projection into the shared space. Alignment is quantified using three complementary metrics: cross-model correlation, root-mean-square (RMS) discrepancy, both computed per embedding dimension across stimuli and then averaged, and cross-model stimulus retrieval accuracy, which measures whether a stimulus in one model maps nearest to the same stimulus in another. See Appendix Section \ref{subsec:metric_desc} for formal definitions and implementation details.

\section{Results}

\subsection{Quantitative evaluation of barycentric alignment}

\begin{figure*}[h!]
\centering
\begin{subfigure}{0.48\linewidth}  
    \centering
    \includegraphics[width=\linewidth]{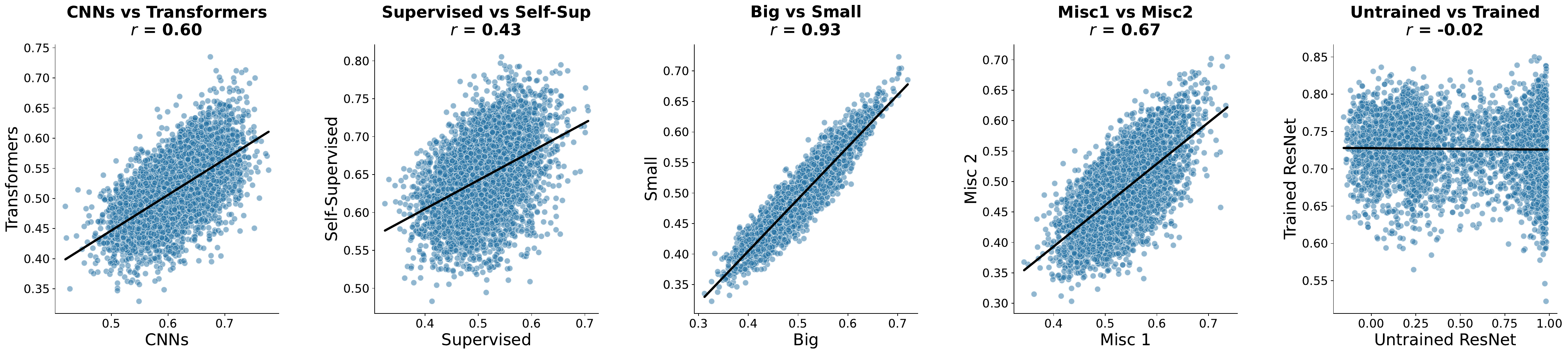}
    \caption{}
\end{subfigure}
\hfill
\begin{subfigure}{0.48\linewidth}  
    \centering
    \includegraphics[width=\linewidth]{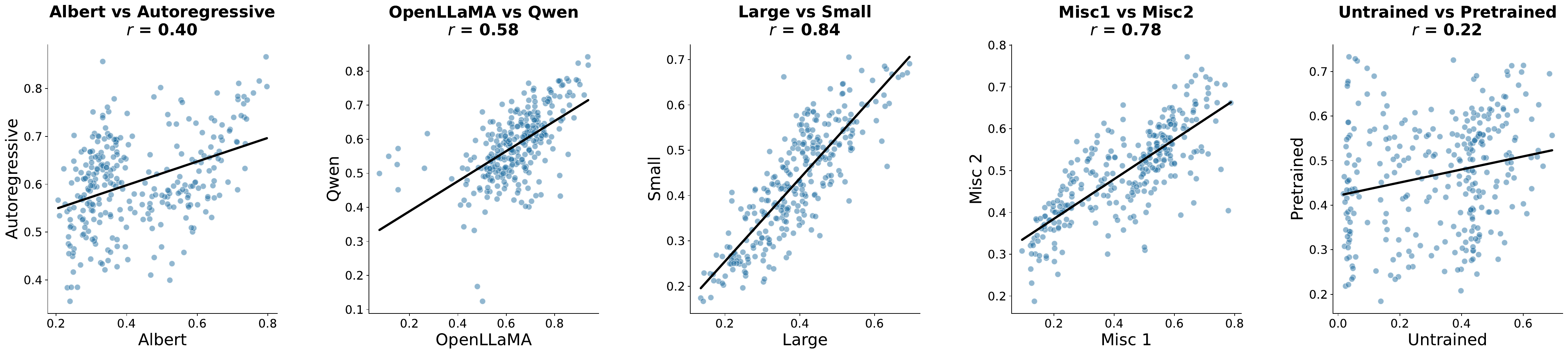}
    \caption{}
\end{subfigure}

\caption{\textbf{Instance-level representational consistency is conserved across model pool constructions.} Pairwise correlations of instance-level representational consistency scores computed from different model pools: (a) vision models and (b) language models, defined by different criteria (architectural family, scale, random mixtures).}
\label{fig:fig1}
\end{figure*}

We first assess the quality of the learned universal embedding space, treating it as a diagnostic of how much representational variation across models can be reduced to nuisance symmetries. Intuitively, if differences between models are largely attributable to symmetry transformations (e.g., rotations of feature space), then barycentric alignment should bring corresponding representations into close correspondence in the shared space. Table~\ref{tab:table1} (top) reports alignment metrics for the supervised vision model pool. Across architectures, models exhibit consistently high correlation in the universal space (often above 0.4), accompanied by low RMS error. Cross-model stimulus retrieval accuracy is also high: top-1 retrieval exceeds 0.64 for all models and approaches 0.9 for top-10 retrieval, despite evaluation over 5,000 held-out stimuli (chance = 0.0002 for top-1; 0.002 for top-10). Language model results are obtained by pooling all LLMs into a single universal space and evaluating retrieval over 300 held-out stimuli (chance top-1 = 0.0033; chance top-10 = 0.033). Alignment remains strong across metrics, with high correlation, low RMS error, and retrieval accuracies far exceeding chance (Table~\ref{tab:table1}, bottom). These results indicate that, after quotienting out orthogonal symmetries, representations from diverse architectures are strongly co-registered at the level of individual stimuli. Results for additional model pools are reported in Appendix Table~\ref{tab:app_fig_1} and show similar trends. Notably, alignment quality is highest for small or homogeneous pools, as expected, but remains strong even for heterogeneous pools spanning architectures, training objectives, and scales. This suggests that a substantial fraction of inter-model variability is symmetry-reducible: once nuisance transformations are accounted for, the underlying representational geometries align closely.

We next examine the instance-level representational consistency scores derived from these aligned spaces and ask whether they depend on how the model pool is constructed. Figure~\ref{fig:fig1}a shows that consistency scores are highly correlated across model pools defined by fundamentally different sources of variability—CNNs versus transformers, supervised versus self-supervised models, large versus small architectures, and even randomly assembled mixed pools. A more detailed comparison across all model pools is provided in Appendix Figure~\ref{fig:app_fig_1}.  Concretely, this means that the same stimuli that expose representational variability across architectures within a given family (e.g., among CNNs) also tend to differentiate models within very different families, such as transformers; likewise, stimuli that separate supervised models also separate self-supervised models, and those that distinguish small models likewise distinguish large ones. This convergence suggests that instance-level disagreement is not tied to any particular mechanism of variability but instead reflects shared axes of representational variation across model classes. Finally, instance-level representational consistency depends critically on learned model structure and is not a trivial property of the input or architectural priors. When the model pool is composed of randomly initialized networks, the resulting instance-level scores are largely uncorrelated with those obtained from trained models (Figure~\ref{fig:fig1}a, Appendix Figure~\ref{fig:app_fig_2}).

For language model pools (Figure~\ref{fig:fig1}b), we observe a qualitatively similar pattern: instance-level representational consistency is broadly shared across different ways of constructing pools, though the strength of agreement varies more than in the vision setting. Correlations are moderate across substantially different architectures and model families and higher across pools with similar composition or size. Interestingly, untrained versus trained comparison shows a weak but still slightly positive correlation, suggesting there’s a small “scaffold” of instance-level structure already present from initialization or tokenization statistics that training partially preserves, even though training still substantially reshapes it.


From a geometric perspective, these results indicate that the \emph{alignability} of representations (how well different systems can be brought into correspondence after quotienting by symmetries) is structured over the input space. Some stimuli occupy stable equivalence classes across models, while others expose irreducible differences in representations.

\subsection{Instance-level representational consistency tracks stimulus difficulty}

\subsubsection{Vision}


\begin{figure}[h!]
\centering
\includegraphics[width=\linewidth]{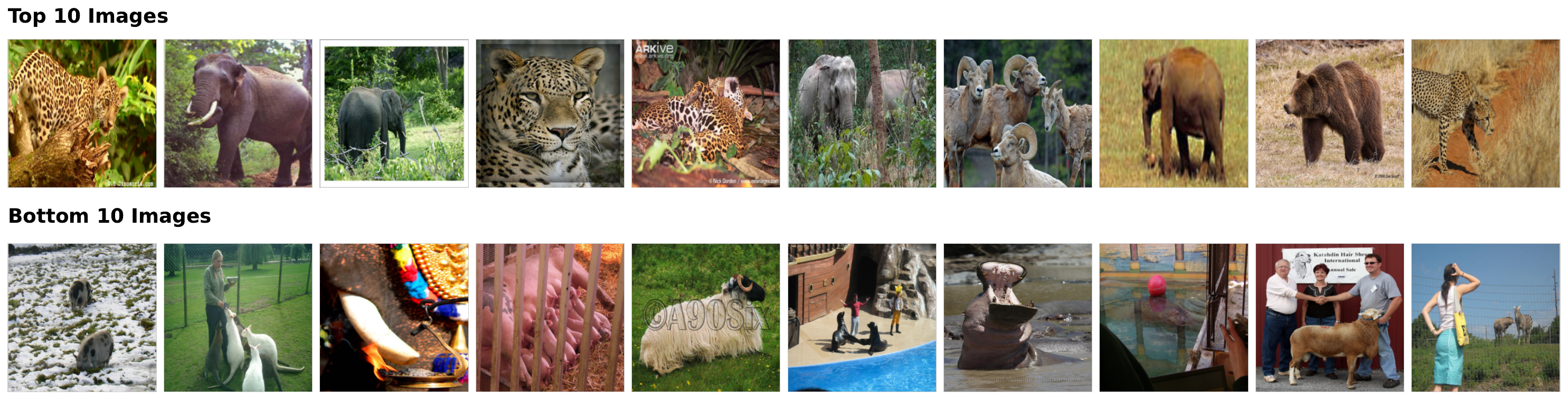}

\vspace{0.3cm} 

\includegraphics[width=0.8\linewidth]{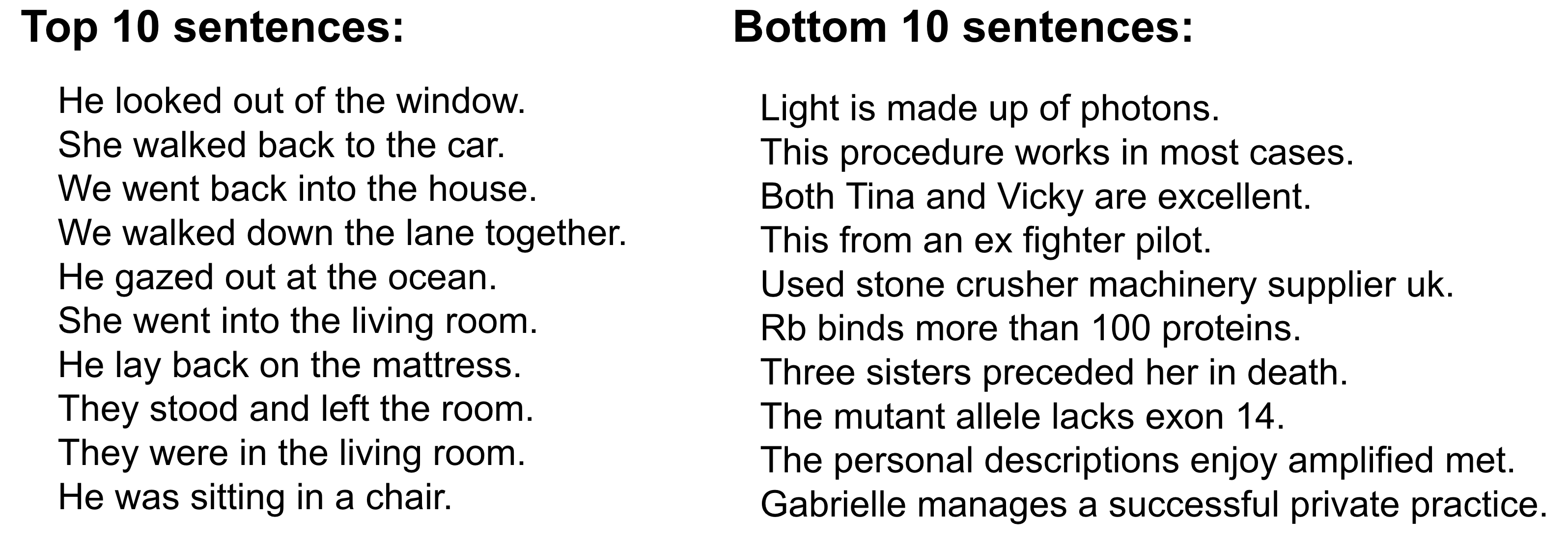}

\caption{Representative images from the Animals parent category with the highest and lowest instance-level representational consistency scores, derived from a pooled ensemble of supervised models with varied architectures and scales.}
\label{fig:fig2}
\end{figure}

\begin{figure}[h!]
\centering
\includegraphics[width=0.9\linewidth]{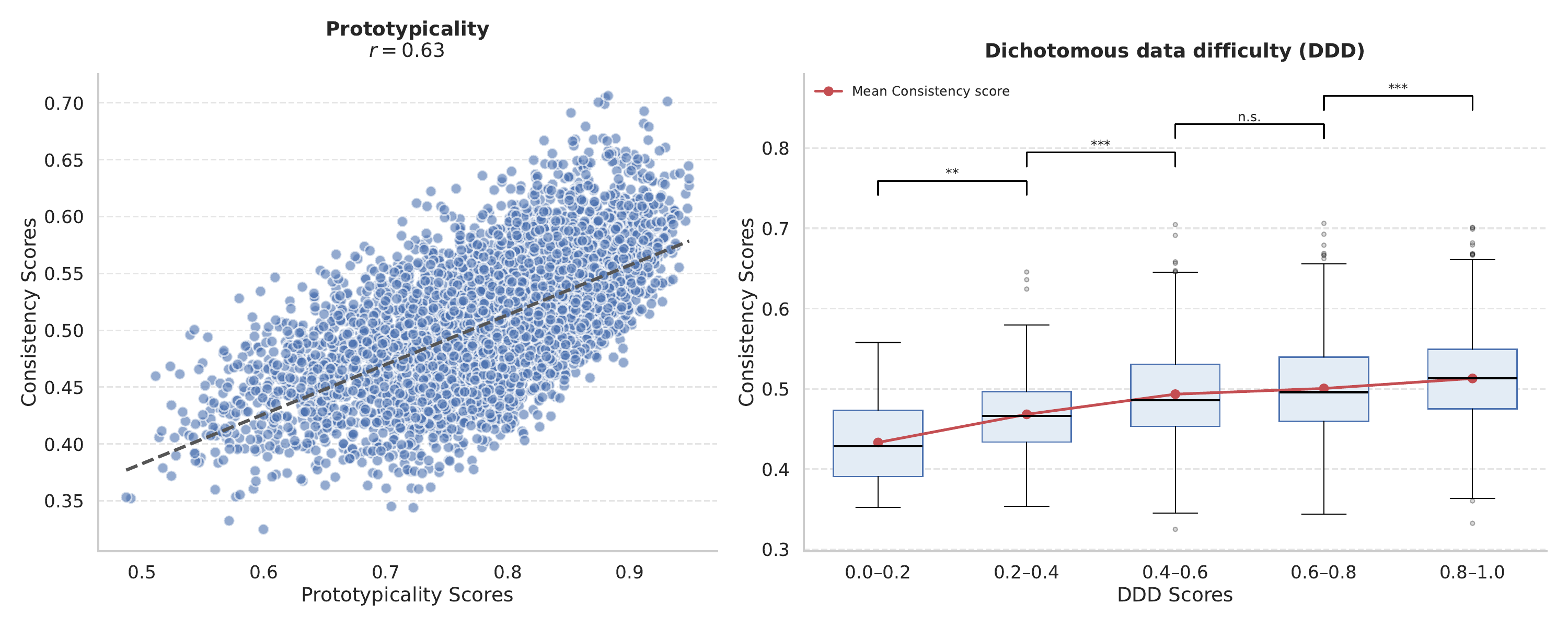}
\caption{Instance-level representational consistency as a function of computational difficulty metrics (Prototypicality and DDD). Paired t-tests across DDD bins show significant increases in consistency scores.}
\label{fig:fig3a}
\end{figure}

\begin{figure}[h!]
\centering
\includegraphics[width=0.9\linewidth]{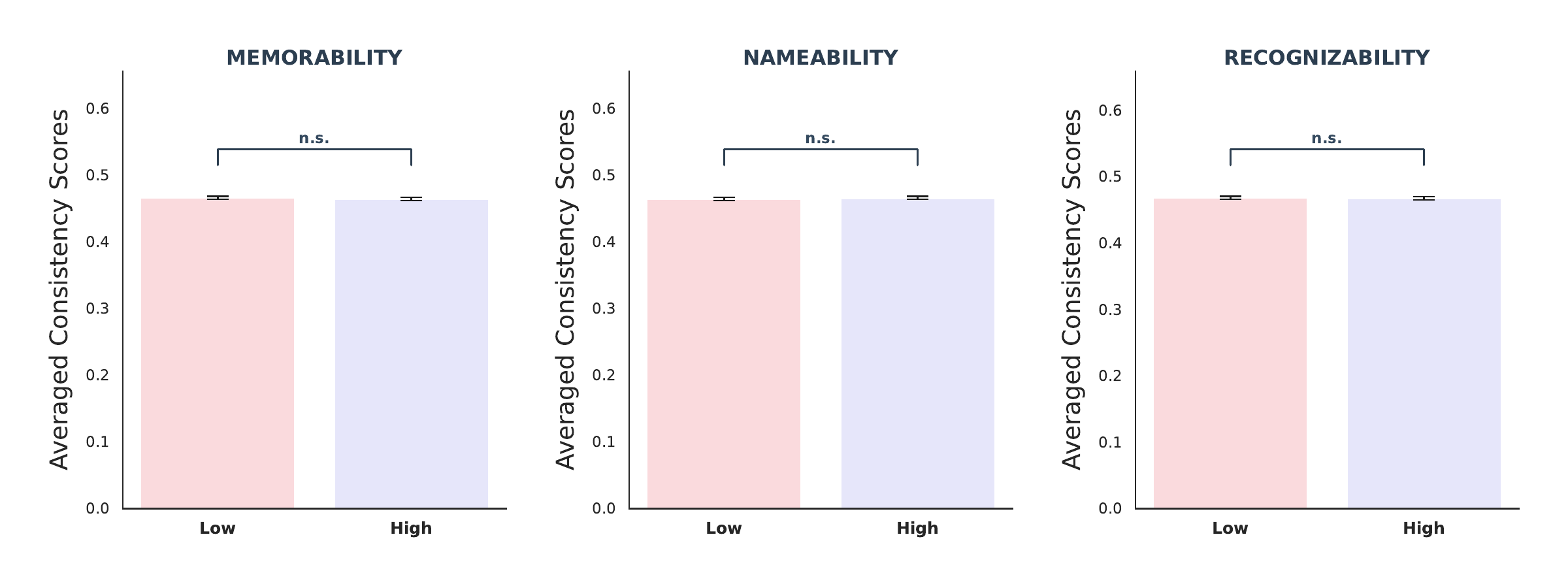}
\caption{Statistical analysis of instance-level representational consistency as a function of human behavioral metrics (memorability, nameability, and recognizability). Paired t-tests indicate that the observed differences are not statistically significant.}
\label{fig:fig3b}
\end{figure}


We next examinine how instance-level representational consistency varies across natural images. Ranking ImageNet validation images by their consistency scores reveals a striking qualitative structure (Figure \ref{fig:fig2}, Appendix Figure \ref{fig:app_fig_7}): images with high consistency are visually unambiguous and strongly aligned with their nominal categories, whereas low-consistency images tend to be perceptually ambiguous, cluttered, or atypical, even for human observers. This qualitative pattern suggests that instance-level representational consistency may reflect stimulus difficulty.

To formalize this relationship, we focus on consistency scores computed from the ``Supervised'' vision model pool and relate them to established computational measures of stimulus difficulty. We first consider \emph{prototypicality} which quantifies how representative an image is of its semantic category, with more prototypical images lying closer to the category centroid in feature space. Instance-level consistency is strongly correlated with prototypicality (Figure~\ref{fig:fig3a}), indicating that images that are more visually canonical also elicit more convergent representations across models. This correspondence suggests that representational consensus concentrates on stimuli that occupy stable, central regions of category structure.

We next relate instance-level representational consistency to \emph{dichotomous data difficulty (DDD)}, an agreement-based measure that quantifies how many models assign the same label to a stimulus, irrespective of correctness. Consistency scores increases systematically with DDD (Figure \ref{fig:fig3a}), demonstrating that stimuli on which models agree at the level of decision outputs are also those on which their internal representations align most closely. This correspondence indicates that instance-level representational consistency captures a shared notion of stimulus difficulty that is reflected both in models’ internal representations and in their downstream decisions, clarifying how population-level behavioral agreement emerges from shared structure in representation space.

Finally, we investigated the relationship between instance-level representational consistency and human perceptual judgments from the THINGS dataset \cite{hebart2023things}, including memorability, nameability, and recognizability. Memorability reflects the likelihood that an image is remembered after brief exposure, nameability quantifies the consistency of object labels assigned by human observers, and recognizability measures how easily an object can be correctly identified. We found that instance-level representational consistency shows no significant relationship with any of these measures. (Figure \ref{fig:fig3b}). Overall, these results suggest that instance-level representational consistency is more closely tied to difficulty as it arises for a population of models than to human perceptual experience.

\subsubsection{Language}

\begin{figure}[h!]
\centering
\includegraphics[width=0.9\linewidth]{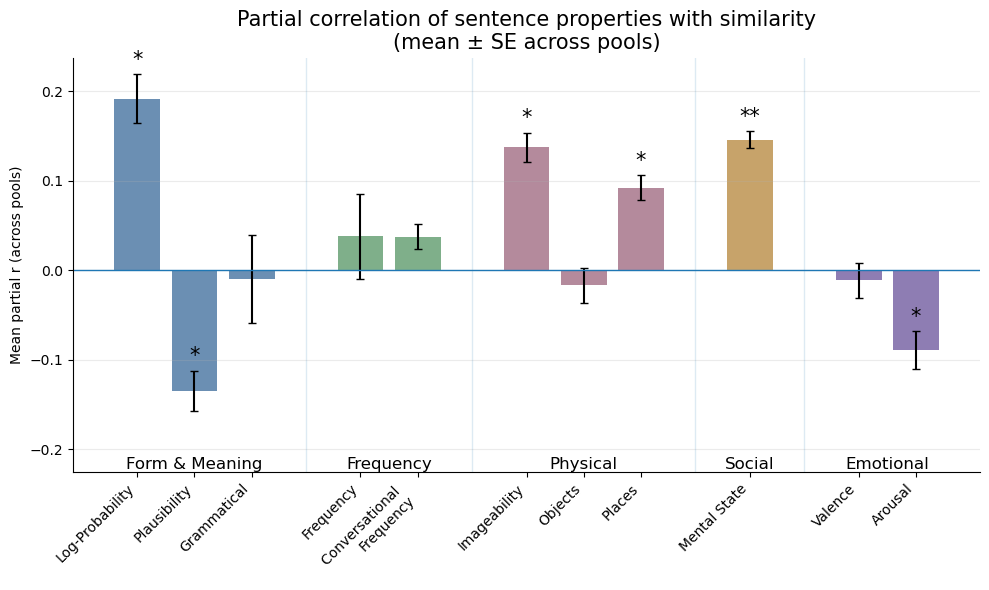}
\caption{Partial correlations between sentence properties and instance-level representational consistency. Bars show the mean partial correlation r across k=4 language-model pools; error bars indicate ±1 SE across pools.}
\label{fig:sentence_prop}
\end{figure}

We next apply the framework to language, examining instance-level representational consistency scores across diverse language model pools using the curated six-word stimulus set described in Section \ref{greta-data}. We compute an instance-level representational consistency score for each sentence and analyze both qualitative exemplars at the extremes and quantitative relationships with annotated linguistic and semantic properties.

Qualitatively, sentences with the highest consistency scores (Figure \ref{fig:fig2}, bottom left) are conventional and scene-based, describing everyday physical actions and locations using simple syntax and high-frequency lexical items (e.g., walked back, went into the living room, sitting in a chair). In contrast, sentences with the lowest consistency scores (Figure \ref{fig:fig2}, bottom right) tend to be stylistically and semantically less homogeneous, including technical/scientific or biomedical statements, fragment-like constructions, and web/SEO-like artifacts, which introduce rarer tokens, abbreviations, and domain-specific terminology.

To characterize this structure more formally, we computed partial correlations between sentence-level consistency and a suite of lexical and human-rated norms (Figure~\ref{fig:sentence_prop}). The strongest independent predictor of consistency is distributional predictability: sentence-level consistency increases with log-probability (mean partial $r=0.19$), indicating that sentences that are more expected under language statistics elicit more convergent representations across model pools even after accounting for frequency and semantic factors. In contrast, frequency exhibits only weak partial associations (general and conversational frequency: mean partial $r\approx 0.04$), suggesting that its marginal correlation with consistency is largely explained by shared variance with predictability.

Beyond predictability, we observe a contribution of concrete/grounded semantic content, with positive partial correlations for imageability (mean partial $r=0.14$) and place-related content (mean partial $r=0.09$). Social-cognitive content also shows a positive association (mental state, mean partial $r=0.15$), whereas emotional dimensions are weak (valence: mean partial $r\approx 0$; arousal: mean partial $r=-0.09$). Interestingly, plausibility shows a negative partial association with consistency (mean partial $r=-0.13$), indicating that after controlling for predictability and other norms, plausibility ratings capture residual variation that is associated with greater cross-model divergence.

Taken together, these results suggest that cross-model representational agreement concentrates most strongly on sentences that are distributionally predictable, with additional contributions from concrete, physically grounded content, while affective dimensions play a comparatively minor role.

\subsection{Cross-modal barycentric alignment}

Recent work on the Platonic Representation Hypothesis \cite{huh2024platonic} has proposed that sufficiently large models trained on different modalities may converge on a shared representational structure, as evidenced by increasing overlap in local neighborhoods across representation spaces as model scale grows. This result suggests a form of cross-modal convergence at scale, but leaves open a critical question: to what extent are differences between vision and language models reducible to symmetry transformations, such that their representations admit a joint alignment to a common reference space?

To address this question, we extend our barycentric alignment framework to a cross-modal setting by jointly aligning a pool of vision-only and language-only models into a shared quotient space defined by a common barycenter. This approach can be contrasted with learned cross-modal models such as CLIP \cite{radford2021learning}, which acquire a shared image–text space through large-scale paired supervision during representation learning. In contrast, our approach does not learn a new cross-modal encoder; instead, it makes explicit how much cross-modal structure is already present in pretrained unimodal representations and can be revealed through symmetry-aware alignment. 


We constructed a joint universal embedding space using a pool of vision transformers (ViTs) and language models (LLMs). For each MS-COCO image–caption pair, we extracted (i) image embeddings from each vision model and (ii) caption embeddings from each language model. We then learned a set of model-specific alignment transforms that map all representations from both modalities into a common universal space (Figure~\ref{fig:multimodal}a). 

\subsubsection{Cross-modal Barycentric Alignment Quality and Retrieval} 

\begin{figure*}[h!]
\centering
\begin{subfigure}[b]{0.3\linewidth}
    \centering
    \includegraphics[width=\linewidth]{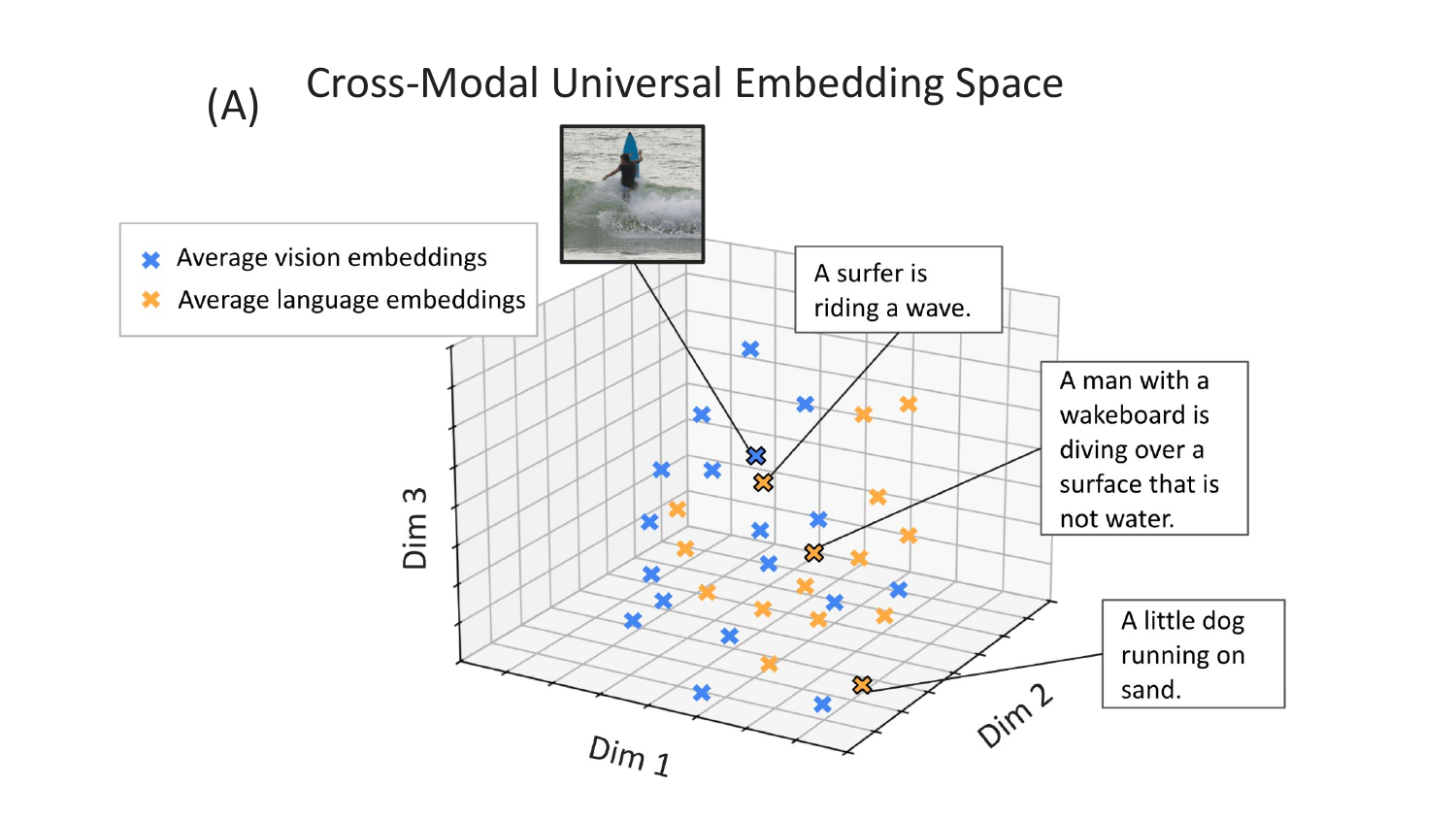}
    \label{fig:a}
\end{subfigure}
\hfill
\begin{subfigure}[b]{0.3\linewidth}
    \centering
    \includegraphics[width=\linewidth]{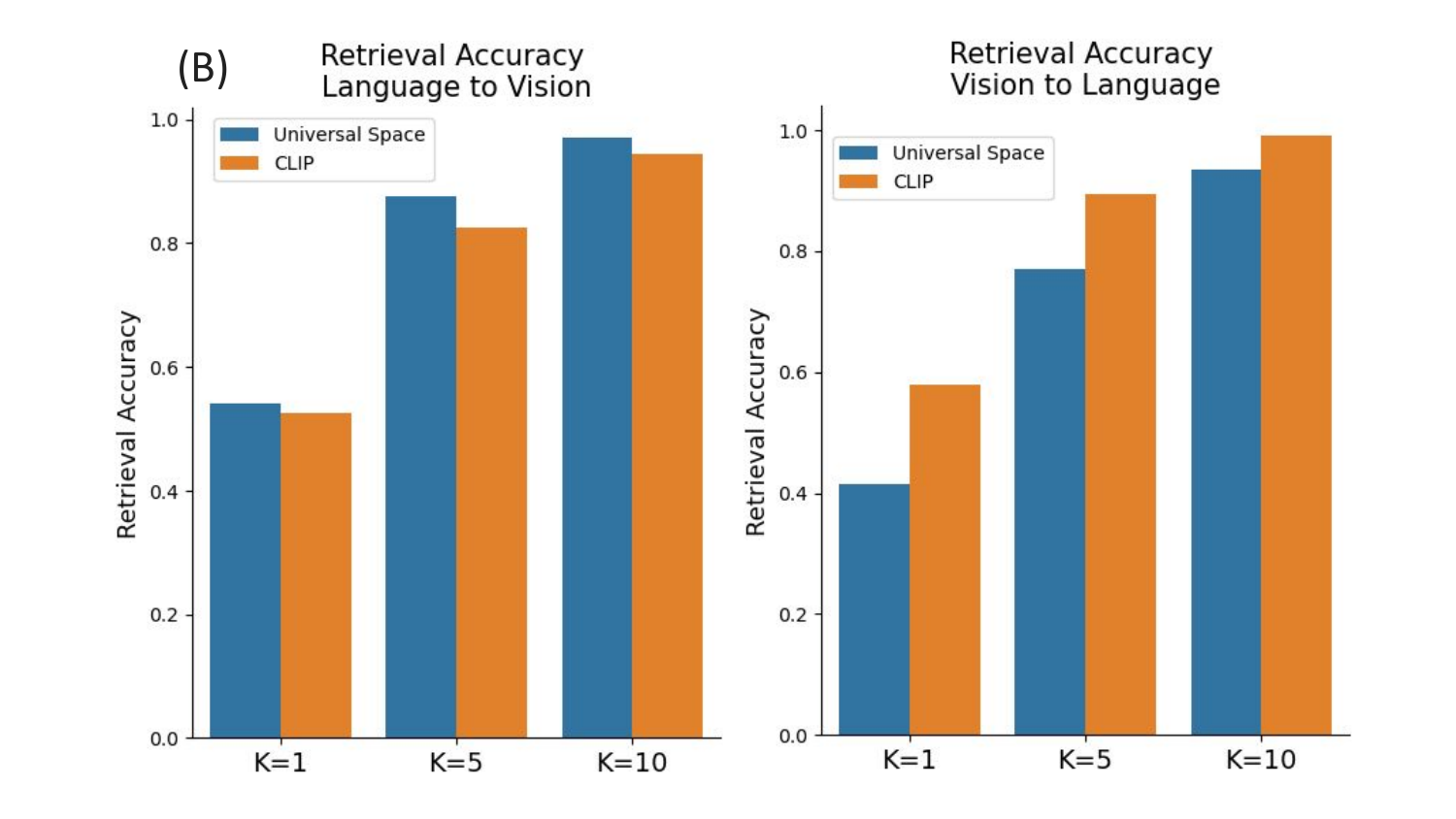}
    \label{fig:b}
\end{subfigure}
\hfill
\begin{subfigure}[b]{0.3\linewidth}
    \centering
    \includegraphics[width=\linewidth]{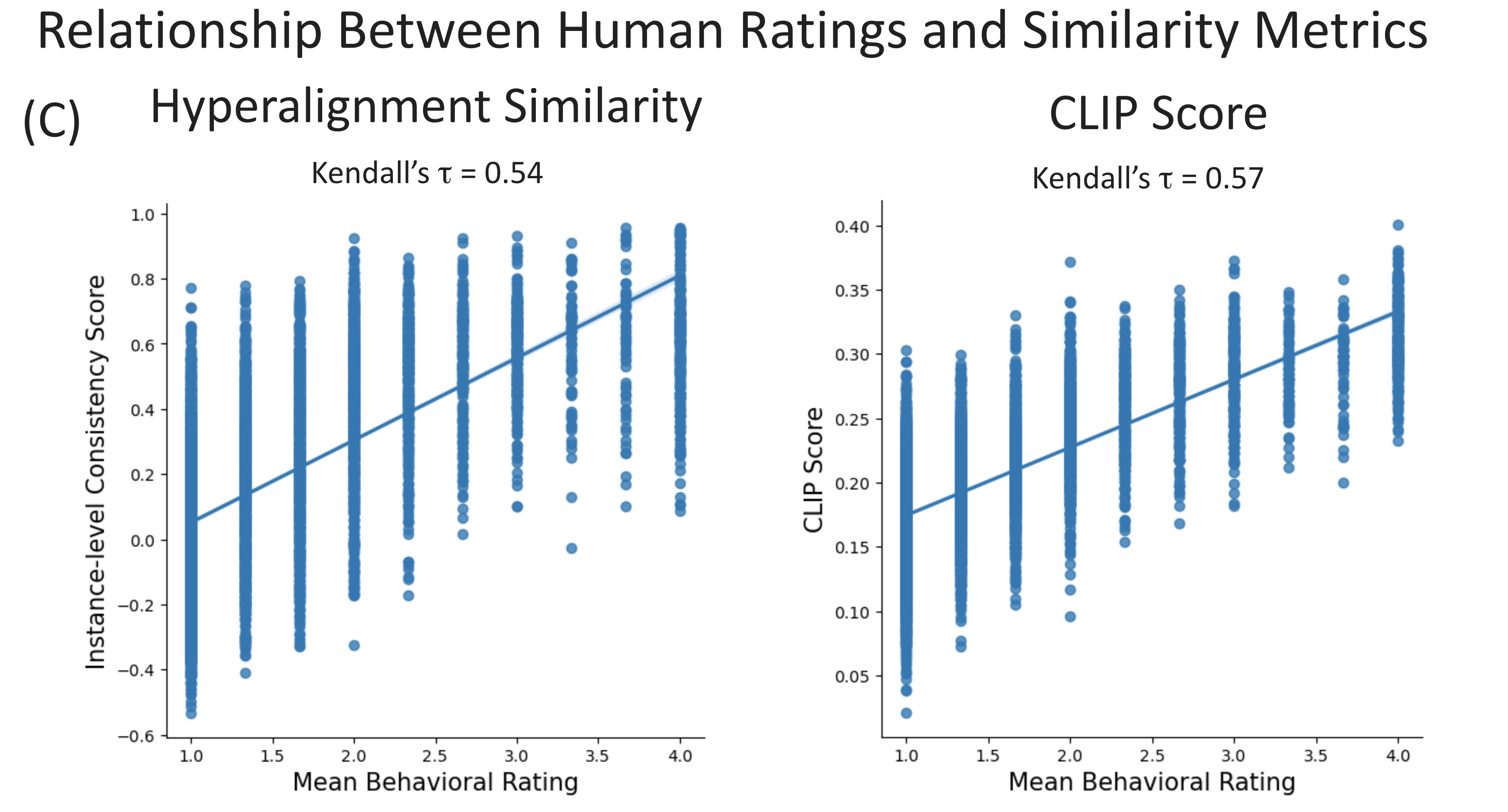}
    \label{fig:c}
\end{subfigure}
\caption{(A) Universal embedding space aligning vision-only image embeddings (blue) and language-only caption embeddings (orange), bringing matched image--caption pairs (e.g., “A surfer is riding a wave”) close together. (B) Bidirectional MS-COCO retrieval accuracy for the universal space and CLIP (text$\rightarrow$image and image$\rightarrow$text; top-$k$, $k\in{1,5,10}$). (C) Flickr8k-Expert comparison: unimodal-pool sentence-level consistency (left) and CLIP image--text similarity (right) versus mean human ratings. Unimodal models used for universal embedding space: OpenLLaMA-7b, OpenLLaMA-13b, ViT-L/14 DINOv2, and ViT-g/14 DINOv2.}
\label{fig:multimodal}
\end{figure*}

We performed a depth-wise analysis of cross-modal retrieval on a held-out test split (n = 200 pairs), with the universal space learned using only 800 training pairs. We evaluated cross-modal retrieval on a held-out test split by measuring bidirectional retrieval accuracy (text$\rightarrow$image and image$\rightarrow$text) at $k\in{1,5,10}$. 

Overall, the universal space yields retrieval performance comparable to CLIP across retrieval depths (Figure~\ref{fig:multimodal}b). In the text$\rightarrow$image direction, our aligned universal space slightly outperforms CLIP at top-1 retrieval (54.0\% vs.\ 52.5\%) while remaining close at top-5 (87.5\% vs.\ 82.5\%) and top-10 (97.0\% vs.\ 94.5\%). In the image$\rightarrow$text direction, CLIP remains higher at each $k$ (top-1: 58.0\% vs.\ 41.5\%; top-5: 89.5\% vs.\ 77.0\%; top-10: 99.0\% vs.\ 93.5\%), but the universal space still achieves strong retrieval accuracy despite relying on pretrained unimodal models and using paired data only to estimate alignment transforms rather than to learn representations.

These findings reveal that unimodal vision and language models already encode, implicitly and independently, much of the structure needed for cross-modal matching. Barycentric alignment exposes this latent compatibility—achieving near-CLIP performance without contrastive pretraining. 



\subsubsection{Cross-Modal Consistency Scores Correlate with Human Judgements} 

To assess whether instance-level representational consistency scores capture human-perceived image–caption matching beyond retrieval, we evaluated the resulting compatibility scores on Flickr8k-Expert, which provides human ratings for image–caption pairs. We computed the similarity between an image and caption as the cosine similarity between their projections in the joint universal space and compared these scores to human judgments. 

In Figure~\ref{fig:multimodal}c, similarities in the universal space were significantly correlated with human ratings (Kendall’s tau = 0.54), closely approaching CLIP’s correspondence with human judgments on the same benchmark (Kendall’s tau = 0.57). While CLIP remains modestly stronger in absolute terms, the gap is small relative to the difference in training regime: CLIP learns a joint space through large-scale paired supervision, whereas the universal embedding is obtained by post hoc alignment of independently trained unimodal models.

Importantly, this result shows that barycentric alignment yields a cross-modal similarity measure that tracks how humans relate meaning across modalities, rather than merely reflecting low-level statistical overlap. That such correspondence emerges without training a dedicated multimodal encoder suggests that substantial alignment-relevant structure is already implicit in unimodal representations and can be made explicit through symmetry-aware alignment.

Taken together, these findings demonstrate that the barycentric alignment framework extends naturally to the cross-modal setting, supporting both paired retrieval and human-aligned compatibility scoring. More broadly, it provides a principled lens for decomposing cross-modal alignment into components that are recoverable via geometry-preserving, post hoc transformations of pretrained representations (a form of shallow alignment) versus components that require contrastive multimodal training and a deeper reshaping of representational geometry.

\subsection{Barycentric alignment of neural representations}

\begin{table}[ht]
\centering
\caption{Average performance across NSD subjects for each region pool (Chance level retrieval scores are 0.01, 0.05 and 0.10 respectively for top-1, top-5 and top-10 respectively.}
\label{tab:table2}
\footnotesize
\addtolength{\tabcolsep}{-2pt} 
\begin{tabular}{@{} l ccccc @{}}
\toprule
\textbf{Region} & \textbf{Corr.} & \textbf{RMS} & \textbf{Top-1} & \textbf{Top-5} & \textbf{Top-10} \\
\midrule
V1v     & 0.277 & 0.048 & 0.243 & 0.458 & 0.579 \\
V1d     & 0.259 & 0.043 & 0.255 & 0.448 & 0.559 \\
V2v     & 0.233 & 0.041 & 0.216 & 0.467 & 0.590 \\
V2d     & 0.209 & 0.050 & 0.179 & 0.374 & 0.505 \\
V3v     & 0.213 & 0.049 & 0.173 & 0.412 & 0.540 \\
V3d     & 0.210 & 0.053 & 0.161 & 0.369 & 0.498 \\
V4      & 0.195 & 0.048 & 0.198 & 0.417 & 0.544 \\
Ventral & 0.159 & 0.016 & 0.203 & 0.503 & 0.668 \\
Lateral & 0.150 & 0.023 & 0.117 & 0.318 & 0.446 \\
Dorsal  & 0.170 & 0.015 & 0.139 & 0.389 & 0.575 \\
\bottomrule
\end{tabular}
\end{table}

\begin{figure}[h!]
\centering
\includegraphics[width=\linewidth, trim=0 0 15.8cm 0, clip]{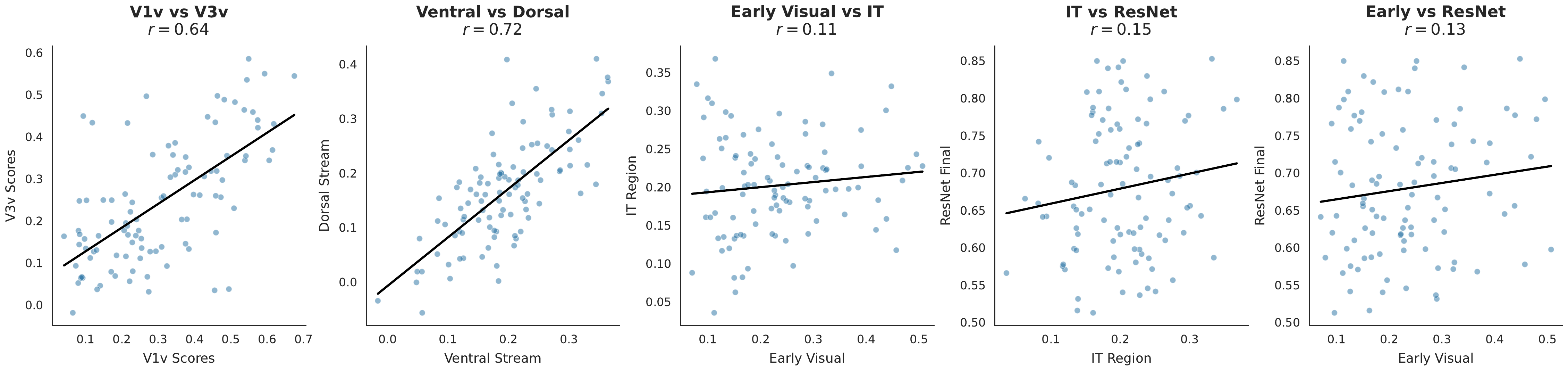}
\caption{Instance-level representational consistency scores computed from different cortical region pools. Scores within early visual region pools show strong consistency with one another, as do scores within higher-order visual region pools. In contrast, correlation between scores between early and higher-order region pools is substantially weaker.}
\label{fig:fig7}
\end{figure}

Next, we extend the universal embedding framework to biological data by constructing pools of neural representations derived from human visual cortex responses from the Natural Scenes Dataset \cite{allen2022massive}(Appendix Table \ref{tab:brain-data-groups}). Each cortical region from each subject is treated as a distinct model. These models are grouped by functional stream (e.g., pooling all ventral visual regions across subjects).

These biological pools span qualitatively distinct functional roles. Early visual cortical regions primarily encode low-level visual features such as edges and orientations, whereas higher-order visual regions support more abstract, image-centric representations that more closely align with the objectives typically optimized in artificial vision models, such as object recognition and segmentation.

We compute instance-level representational consistency scores using images from the MS COCO dataset that were included in the Natural Scenes Dataset \cite{allen2022massive}. Of these, 1,000 images were viewed by four subjects, and we construct a train–test split over this shared stimulus set, evaluating all metrics on held-out images.

We first assess the quality of the resulting universal embedding using the same metrics applied to artificial systems. Table \ref{tab:table2} reports these metrics for pools drawn from each individual region in the visual cortex. For each region, scores are averaged across the 4 subjects (Appendix Table \ref{tab:model_performance_avg} reports scores for individual subjects). Despite the substantial noise inherent in fMRI-derived representations, correlation scores remain relatively high (exceeding 0.2 without noise normalization), RMS errors are low, and retrieval accuracies are consistently above chance. These demonstrate that barycentric alignment yields a stable and meaningful shared space even when applied to noisy biological population codes. 

We next examine whether instance-level representational consistency scores are preserved across different biological pools. Strikingly, scores computed within pools drawn from comparable cortical regions are highly correlated: early visual region pools exhibit strong mutual agreement with one another, as do pools drawn from higher-order visual areas. In contrast, correlations between early and higher visual pools are substantially weaker (Figure \ref{fig:fig7}, Appendix Figure \ref{fig:app_fig_9}). This pattern mirrors the known stratification of the visual hierarchy and indicates that instance-level representational agreement is structured by cortical stage rather than reflecting a generic property of the stimulus set.

Notably, instance-level representational consistency scores computed from biological model pools show little correlation with those obtained from artificial model pools, such as the ResNet family (Figure \ref{fig:fig7}, Appendix Figure \ref{fig:app_fig_10}). This dissociation indicates that the stimuli on which biological models exhibit strong consensus differ markedly from those that elicit agreement among artificial vision models. Strikingly, this divergence persists even for the IT (Later) model pool, which comprises high-level visual cortical regions whose functional role most closely parallels the objectives optimized in artificial networks, including object classification. Together, these results suggest that the dominant sources of variability captured by contemporary artificial model pools (e.g. architecture) do not yet provide a sufficient basis for explaining the structure of representational variability observed across biological individuals.

\section{Discussion}
We proposed a simple barycentric alignment framework for aligning \emph{populations} of representation spaces into a common universal embedding, quotienting out nuisance symmetries. The key advantage is granularity: once models (individuals/regions) are co-registered in a shared quotient space, representational agreement can be defined at the \emph{instance level} rather than only as a set-level summary, yielding a direct way to identify which inputs expose stable versus variable representations across populations spanning architectures, objectives, scales, modalities, and biological groups.

Across vision and language model pools, instance-level consistency is highly conserved across many ways of constructing the population and varies systematically with stimulus properties, yet it dissociates sharply between trained and untrained networks, indicating that the agreement landscape is shaped by learning rather than architectural priors. Extending the same framework across modalities, we find that post-hoc, symmetry-aware alignment of unimodal vision and language models yields image–text similarity scores that closely track human judgments of cross-modal compatibility, demonstrating that how humans relate meaning across modalities is already partially encoded in unimodal representations and can be recovered through shallow post-hoc alignment. Finally, applying the method to fMRI-derived representations yields stable alignment within biological pools but weak correspondence with artificial pools, suggesting that the dominant sources of variability probed by current model ensembles (e.g., architecture) may be insufficient to capture inter-individual biological variation. More broadly, the barycentric perspective opens several directions for future work, including distinguishing cross-modal structure that is recoverable via geometry-preserving alignment from structure that requires joint multimodal training, identifying artificial model populations whose variability structure better matches inter-individual biological variability, and characterizing when quotient-based universal spaces generalize beyond the training domain.

\nocite{langley00}

\section*{Impact Statement}

This paper presents work whose goal is to advance the field of Machine Learning. There are many potential societal consequences of our work, none of which we feel must be specifically highlighted here.

\bibliography{example_paper}
\bibliographystyle{icml2026}

\clearpage
\appendix
\renewcommand{\thesection}{\Alph{section}}
\setcounter{section}{0}
\setcounter{figure}{0}
\setcounter{table}{0}
\setcounter{equation}{0}

\twocolumn[
\section{Appendix}
]

\tcbset{
    modernalg/.style={
        enhanced,
        colback=boxBack,
        boxrule=1pt,
        arc=4mm,
        fonttitle=\sffamily\bfseries,
        left=3mm,
        right=3mm,
        top=2mm,
        bottom=2mm,
        shadow={1mm}{-1mm}{0mm}{black!15},
        boxsep=1mm
    }
}

\begin{tcolorbox}[modernalg, title=Phase I: Training (Barycenter), colframe=algBlue, coltitle=white, label={alg:procrustes-barycenter-training}]
\begin{algorithm}[H]
    \SetAlgoLined
    \SetKwInOut{Input}{Input}
    \SetKwInOut{Output}{Output}
    
    \Input{$X_i \in \mathbb{R}^{n \times d}, T, \varepsilon$}
    \Output{$M^{(t+1)}, \{T_i\}$}
    
    \BlankLine
    Initialize $M^{(0)} \gets \frac{1}{N}\sum_i X_i$\;
    
    \For{$t = 0$ \KwTo $T-1$}{
        \For{$i = 1$ \KwTo $N$}{
            $T_i \gets \arg\min_{R \in O(d)} \|X_i R - M^{(t)}\|_F$\;
            $X_i^{(t)} \gets X_i T_i$\;
        }
        $M^{(t+1)} \gets \frac{1}{N}\sum_{i=1}^N X_i^{(t)}$\;
        
        \If{$\frac{\|M^{(t+1)} - M^{(t)}\|_F}{\|M^{(t)}\|_F} < \varepsilon$}{
            \textbf{break}\;
        }
    }
\end{algorithm}
\end{tcolorbox}

\begin{tcolorbox}[modernalg, title=Phase II: Inference (Scoring), colframe=algOrange, coltitle=white, label={alg:procrustes-barycenter-inference}]
\begin{algorithm}[H]
    \SetAlgoLined
    \SetKwInOut{Input}{Input}
    \SetKwInOut{Output}{Output}
    
    \Input{$Y_i \in \mathbb{R}^{m \times d}, \{T_i\}$}
    \Output{$\{S_1, \dots, S_m\}$}
    
    \BlankLine
    \For{$i = 1$ \KwTo $N$}{
        $Y_i' \gets Y_i T_i$\;
    }
    
    \BlankLine
    \For{$j = 1$ \KwTo $m$}{
        $S_j \gets \frac{1}{N(N-1)} \sum_{p \neq q}^{N} \text{SIM}(\mathbf{Y}_{pj}', \mathbf{Y}_{qj}')$\;
    }
\end{algorithm}
\end{tcolorbox}


\definecolor{tableblue}{RGB}{245, 247, 251}

\begin{table}[t]
\centering
\footnotesize 
\renewcommand{\arraystretch}{1.2}
\setlength{\tabcolsep}{4pt} 
\caption{\textbf{Vision model pools.} Categorized by architecture, scale, and training objective for Procrustes alignment.}
\label{tab:model-pools}

\begin{tabularx}{\columnwidth}{@{} l >{\RaggedRight\arraybackslash}X @{}} 
\toprule
\textbf{Group} & \textbf{Models} \\
\midrule

\rowcolor{tableblue} \multicolumn{2}{l}{\textbf{\textit{Architecture-based pools}}} \\ 
ResNets      & \texttt{resnet18, resnet34, resnet50, resnet101, resnet152} \\
VGGs         & \texttt{vgg11, vgg13, vgg16, vgg19} \\
ConvNeXts    & \texttt{convnext\_tiny, convnext\_small, convnext\_base, convnext\_large} \\
ViTs         & \texttt{vit\_t\_16, vit\_s\_16, vit\_b\_16, vit\_l\_16} \\
DINO ViTs    & \texttt{vit\_s\_16\_dino, vit\_b\_16, vit\_b\_8} \\
Swins        & \texttt{swin\_t, swin\_s, swin\_b} \\

\addlinespace[0.4em]

\rowcolor{tableblue} \multicolumn{2}{l}{\textbf{\textit{Scale and objective-based pools}}} \\
Large        & \texttt{resnet101, resnet152, vgg16, vgg19, swin\_b, vit\_b\_16, convnext\_base, convnext\_large} \\
Small        & \texttt{resnet18, resnet34, resnet50, vgg11, vgg13, swin\_t, swin\_s, vit\_s\_16, vit\_t\_16} \\
Supervised   & \textit{Standard variants of ResNet, VGG, ViT, Swin, and ConvNeXt.} \\
Self-sup.    & \texttt{resnet50\_dino, moco\_v2, vit\_s\_16\_dino, vit\_b\_16\_dino} \\

\addlinespace[0.4em]

\rowcolor{tableblue} \multicolumn{2}{l}{\textbf{\textit{Auxiliary mixed pools}}} \\
Misc-1       & \texttt{resnet34, resnet50\_dino, swin\_b} \\
Misc-2       & \texttt{vgg13, moco\_v2, vit\_s\_16\_dino} \\
Misc-3       & \texttt{alexnet, vit\_s\_16\_dino, swin\_s} \\

\bottomrule
\end{tabularx}
\end{table}

\begin{table*}[t]
\centering
\caption{Correlation score, RMS score, and retrieval accuracies across different vision model pools.}
\label{tab:app_fig_1}
\small
\resizebox{\textwidth}{!}{
\begin{tabular}{l S[table-format=1.4] S[table-format=1.5] S[table-format=1.4] S[table-format=1.4] S[table-format=1.4]}
\toprule
\textbf{Model} & {\textbf{Corr. $\uparrow$}} & {\textbf{RMS $\downarrow$}} & {\textbf{Top-1 $\uparrow$}} & {\textbf{Top-5 $\uparrow$}} & {\textbf{Top-10 $\uparrow$}} \\
\midrule
\multicolumn{6}{c}{\textit{Self Supervised Model Pool}} \\
\midrule
Resnet 50 Dino & 0.5726 & 0.0187 & 0.9889 & 0.9985 & 0.9995 \\
Resnet 50 Moco v2 & 0.5299 & 0.0197 & 0.9760 & 0.9972 & 0.9990 \\
Vit B 16 Dino & 0.6429 & 0.0166 & 0.9669 & 0.9934 & 0.9980 \\
Vit S 16 Dino & 0.6199 & 0.0170 & 0.9732 & 0.9952 & 0.9985 \\
Vit B 8 Dino & 0.6198 & 0.0172 & 0.9647 & 0.9941 & 0.9981 \\
\midrule
\multicolumn{6}{c}{\textit{ResNet Model Pool}} \\
\midrule
Resnet 34 & 0.6959 & 0.0163 & 0.9964 & 0.9999 & 1.0000 \\
Resnet 50 & 0.7270 & 0.0154 & 0.9957 & 1.0000 & 1.0000 \\
Resnet 18 & 0.6775 & 0.0167 & 0.9928 & 1.0000 & 1.0000 \\
Resnet 101 & 0.7252 & 0.0154 & 0.9915 & 0.9999 & 1.0000 \\
Resnet 152 & 0.7239 & 0.0155 & 0.9880 & 0.9998 & 1.0000 \\
\midrule
\multicolumn{6}{c}{\textit{ConvNext Model Pool}} \\
\midrule
Convnext tiny & 0.6325 & 0.0193 & 0.9513 & 0.9974 & 0.9989 \\
Convnext small & 0.6947 & 0.0178 & 0.9607 & 0.9973 & 0.9991 \\
Convnext base & 0.7059 & 0.0173 & 0.9595 & 0.9981 & 0.9997 \\
Convnext large & 0.6964 & 0.0178 & 0.9518 & 0.9977 & 0.9999 \\
\midrule
\multicolumn{6}{c}{\textit{Swin Transformers Model Pool}} \\
\midrule
swin T & 0.7403 & 0.0221 & 0.9858 & 0.9985 & 0.9998 \\
swin B & 0.7682 & 0.0209 & 0.9675 & 0.9987 & 1.0000 \\
swin S & 0.7690 & 0.0208 & 0.9822 & 0.9989 & 1.0000 \\
\midrule
\multicolumn{6}{c}{\textit{Vision Transformers Model Pool}} \\
\midrule
vit B 16 & 0.4243 & 0.0309 & 0.8593 & 0.9595 & 0.9768 \\
vit S 16 & 0.4384 & 0.0294 & 0.9147 & 0.9823 & 0.9907 \\
vit T 16 & 0.3360 & 0.0316 & 0.7233 & 0.9024 & 0.9379 \\
vit L 16 & 0.4287 & 0.0298 & 0.7850 & 0.9215 & 0.9481 \\
\midrule
\multicolumn{6}{c}{\textit{VGG Model Pool}} \\
\midrule
VGG 11 & 0.6768 & 0.0125 & 0.9948 & 1.0000 & 1.0000 \\
VGG 13 & 0.6779 & 0.0125 & 0.9949 & 1.0000 & 1.0000 \\
VGG 16 & 0.6843 & 0.0123 & 0.9893 & 0.9999 & 1.0000 \\
VGG 19 & 0.6790 & 0.0124 & 0.9805 & 0.9998 & 1.0000 \\
\midrule
\multicolumn{6}{c}{\textit{Supervised CNNS Model Pool}} \\
\midrule
resnet 34 & 0.5973 & 0.0130 & 0.9653 & 0.9966 & 0.9992 \\
resnet 50 & 0.6332 & 0.0124 & 0.9623 & 0.9971 & 0.9991 \\
resnet 18 & 0.5960 & 0.0131 & 0.9762 & 0.9985 & 0.9997 \\
resnet 101 & 0.6265 & 0.0125 & 0.9439 & 0.9943 & 0.9982 \\
resnet 152 & 0.6251 & 0.0126 & 0.9334 & 0.9937 & 0.9980 \\
VGG 11 & 0.5894 & 0.0134 & 0.9743 & 0.9990 & 0.9998 \\
VGG 13 & 0.5829 & 0.0135 & 0.9673 & 0.9985 & 0.9997 \\
VGG 16 & 0.5891 & 0.0134 & 0.9543 & 0.9968 & 0.9993 \\
VGG 19 & 0.5885 & 0.0134 & 0.9399 & 0.9958 & 0.9986 \\
alexnet & 0.4686 & 0.0155 & 0.9383 & 0.9938 & 0.9978 \\
\bottomrule
\end{tabular}
\quad \hfill \quad
%
\begin{tabular}{l S[table-format=1.4] S[table-format=1.5] S[table-format=1.4] S[table-format=1.4] S[table-format=1.4]}
\toprule
\textbf{Model} & {\textbf{Corr. $\uparrow$}} & {\textbf{RMS $\downarrow$}} & {\textbf{Top-1 $\uparrow$}} & {\textbf{Top-5 $\uparrow$}} & {\textbf{Top-10 $\uparrow$}} \\
\midrule
\multicolumn{6}{c}{\textit{Supervised Transformers Model Pool}} \\
\midrule
vit B 16 & 0.5059 & 0.0298 & 0.8810 & 0.9730 & 0.9876 \\
vit S 16 & 0.4528 & 0.0309 & 0.8858 & 0.9816 & 0.9927 \\
vit T 16 & 0.3257 & 0.0340 & 0.6342 & 0.8538 & 0.9087 \\
vit L 16 & 0.4802 & 0.0301 & 0.8074 & 0.9519 & 0.9728 \\
swin T & 0.5505 & 0.0283 & 0.8853 & 0.9726 & 0.9866 \\
swin B & 0.5604 & 0.0279 & 0.7905 & 0.9468 & 0.9744 \\
swin S & 0.5571 & 0.0279 & 0.8142 & 0.9549 & 0.9787 \\
\midrule
\multicolumn{6}{c}{\textit{Big Model Pool}} \\
\midrule
vit B 16 & 0.5059 & 0.0298 & 0.8810 & 0.9730 & 0.9876 \\
vit S 16 & 0.4528 & 0.0309 & 0.8858 & 0.9816 & 0.9927 \\
vit T 16 & 0.3257 & 0.0340 & 0.6342 & 0.8538 & 0.9087 \\
vit L 16 & 0.4802 & 0.0301 & 0.8074 & 0.9519 & 0.9728 \\
swin T & 0.5505 & 0.0283 & 0.8853 & 0.9726 & 0.9866 \\
swin B & 0.5604 & 0.0279 & 0.7905 & 0.9468 & 0.9744 \\
swin S & 0.5571 & 0.0279 & 0.8142 & 0.9549 & 0.9787 \\
\midrule
\multicolumn{6}{c}{\textit{Small Model Pool}} \\
\midrule
resnet34 & 0.5171 & 0.0125 & 0.8954 & 0.9785 & 0.9900 \\
resnet50 & 0.5380 & 0.0123 & 0.9110 & 0.9843 & 0.9929 \\
resnet18 & 0.5069 & 0.0127 & 0.8805 & 0.9695 & 0.9849 \\
vgg11 & 0.4390 & 0.0142 & 0.8687 & 0.9761 & 0.9898 \\
vgg13 & 0.4351 & 0.0142 & 0.8649 & 0.9766 & 0.9901 \\
swin T & 0.4908 & 0.0129 & 0.8529 & 0.9770 & 0.9909 \\
swin S & 0.4743 & 0.0131 & 0.7381 & 0.9527 & 0.9820 \\
vit S 16 & 0.4591 & 0.0132 & 0.8390 & 0.9711 & 0.9878 \\
vit T 16 & 0.4001 & 0.0139 & 0.7027 & 0.8748 & 0.9169 \\
convnext tiny & 0.4285 & 0.0134 & 0.7537 & 0.9384 & 0.9670 \\
convnext small & 0.4704 & 0.0129 & 0.7363 & 0.9434 & 0.9713 \\
\midrule
\multicolumn{6}{c}{\textit{Misc1 Model Pool}} \\
\midrule
resnet34 & 0.4905 & 0.0197 & 0.9217 & 0.9892 & 0.9965 \\
Resnet 50 Dino & 0.4855 & 0.0204 & 0.9442 & 0.9915 & 0.9971 \\
swin B & 0.4380 & 0.0212 & 0.7228 & 0.9673 & 0.9918 \\
vit B 16 dino & 0.5463 & 0.0186 & 0.9425 & 0.9923 & 0.9982 \\
\midrule
\multicolumn{6}{c}{\textit{Misc2 Model Pool}} \\
\midrule
vgg13 & 0.4401 & 0.0148 & 0.8651 & 0.9721 & 0.9873 \\
Resnet 50 Moco v2 & 0.4199 & 0.0147 & 0.8631 & 0.9671 & 0.9851 \\
vit S 16 dino & 0.4929 & 0.0135 & 0.9249 & 0.9893 & 0.9973 \\
swin T & 0.4234 & 0.0146 & 0.7706 & 0.9539 & 0.9847 \\
\midrule
\multicolumn{6}{c}{\textit{Misc3 Model Pool}} \\
\midrule
alexnet & 0.3465 & 0.0156 & 0.7991 & 0.9412 & 0.9706 \\
vit S 16 dino & 0.4227 & 0.0137 & 0.8632 & 0.9777 & 0.9928 \\
swin S & 0.3654 & 0.0145 & 0.6408 & 0.9059 & 0.9633 \\
\bottomrule
\end{tabular}}
\end{table*}

\begin{figure}[h!]
\centering
\includegraphics[width=\linewidth]{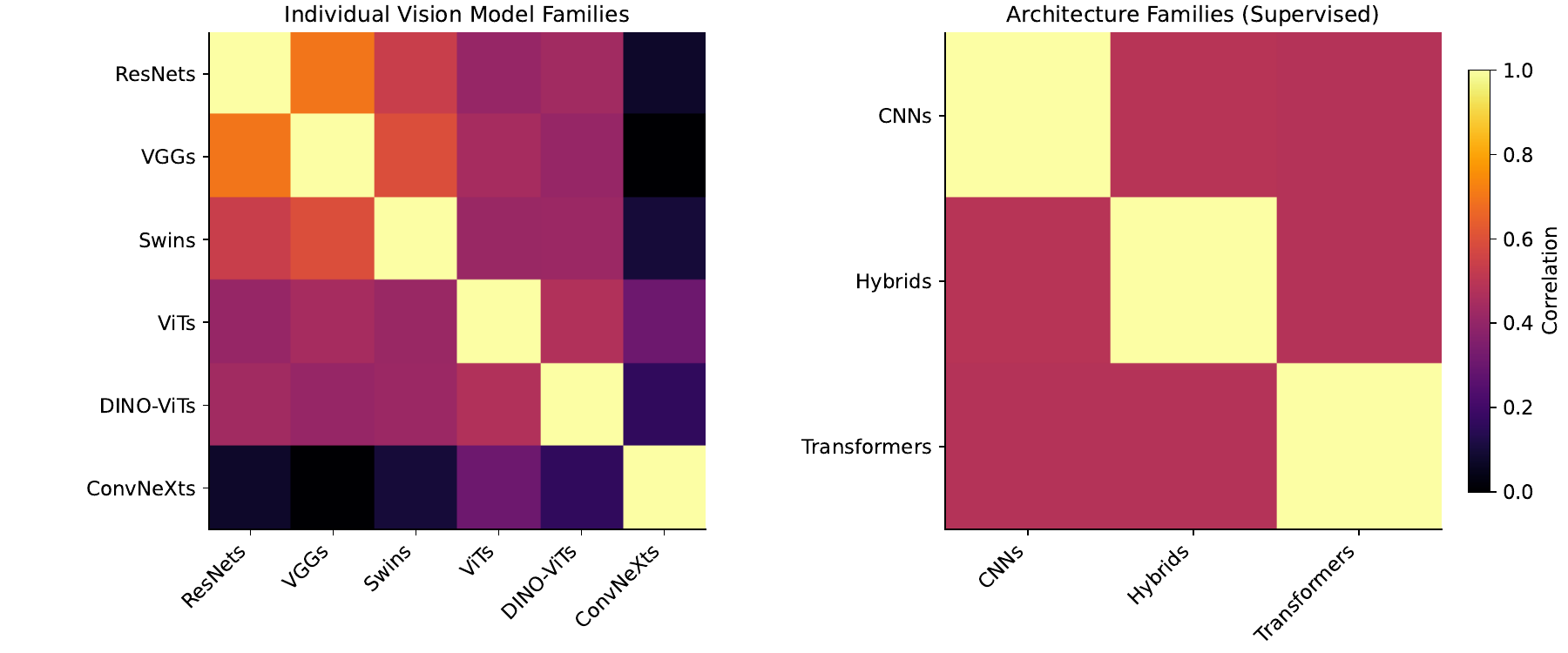}
\caption{Heatmap showing consistency of universal similarity scores across different model families (left), and across different architectures (right)}
\label{fig:app_fig_1}
\end{figure}

\begin{figure}[h!]
\centering
\includegraphics[width=\linewidth]{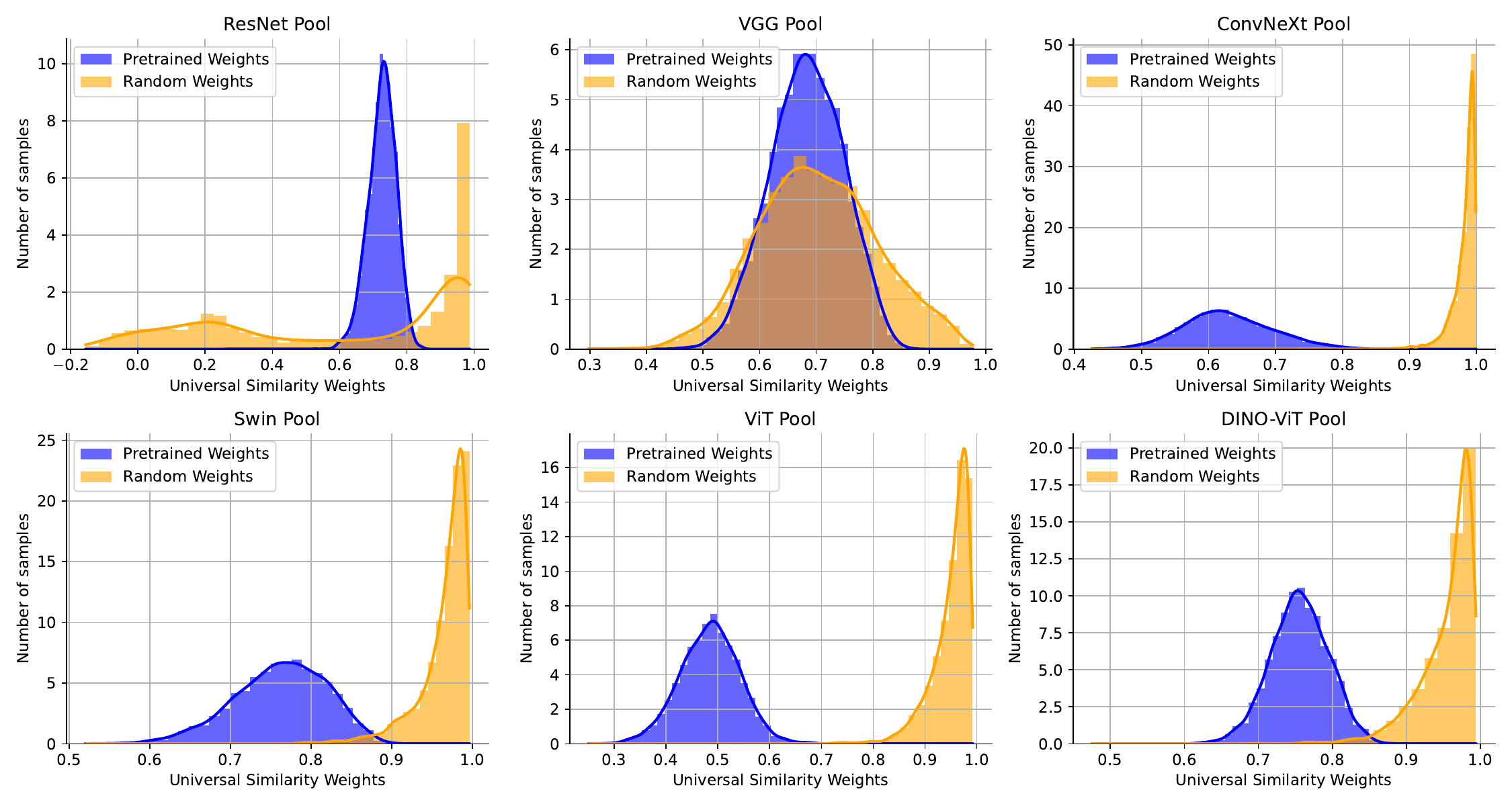}
\caption{Universal Similarity scores depend strongly on the model weights used in the model pool, with substantial differences observed between models with random weights and those with trained weights.}
\label{fig:app_fig_2}
\end{figure}

\subsection{Metrics for evaluating Barycentric Alignment}
\label{subsec:metric_desc}

We quantify the \emph{quality of the learned universal embedding itself} by measuring how well representations from different models align once mapped into the shared space.
Let $\{M_i\}_{i=1}^X$ denote a pool of $X$ models evaluated on the same set of $N$ stimuli. Model $i$ produces a representation matrix $M_i \in \mathbb{R}^{N \times D_i}$. After projection into the universal space, we obtain $M_i' \in \mathbb{R}^{N \times D}$, where $D$ is the shared dimensionality.

\textbf{Correlation Score.}  
For each model pair $(i,j)$ with $i \neq j$ and each universal dimension $d \in \{1,\dots,D\}$, we compute the Pearson correlation across stimuli: $\rho_d^{(i,j)} \;=\; \mathrm{corr}\!\left(M_i'[:,d],\, M_j'[:,d]\right)$.
We report a per-model summary by averaging across dimensions and across all other models: $\mathrm{Corr}(i) \;=\; \frac{1}{(X-1)D}\sum_{\substack{j=1\\ j\neq i}}^{X}\sum_{d=1}^{D}\rho_d^{(i,j)}$.
High values indicate that corresponding universal axes capture similar stimulus-wise variation across models.
 
\textbf{Root Mean Square (RMS) Score.}  
For each pair $(i,j)$ and dimension $d$, we compute the RMS discrepancy across stimuli: $\mathrm{RMS}_d^{(i,j)} \;=\; \sqrt{\frac{1}{N}\sum_{n=1}^{N}\!\left(M_i'[n,d]-M_j'[n,d]\right)^2}$.
The per-model summary averages across dimensions and partners: $\mathrm{RMS}(i) \;=\; \frac{1}{(X-1)D}\sum_{\substack{j=1\\ j\neq i}}^{X}\sum_{d=1}^{D}\mathrm{RMS}_d^{(i,j)}$.
Low values indicate tight co-registration in the shared coordinates.

\textbf{Cross-model Stimulus Retrieval Accuracy.}  
Cross-model stimulus retrieval operationalizes a strong notion of representational agreement: whether different models not only encode stimuli similarly in aggregate, but identify the same stimulus as corresponding to itself across representational spaces. 
For a stimulus $x \in \{1, \dots, N\}$ and a model pair $(M_i, M_j)$ with $i \neq j$, we compute the Euclidean distances $d_{x,y}^{(i,j)} = \left\lVert M_i'[x] - M_j'[y] \right\rVert_2, \quad \forall y \in \{1, \dots, N\}$.
Let $\mathcal{N}_K^{(i,j)}(x)$ denote the indices of the $K$ smallest distances. The retrieval accuracy for stimulus $x$ is $\mathrm{Acc}_K^{(i,j)}(x) = \mathbb{I}\!\left[x \in \mathcal{N}_K^{(i,j)}(x)\right]$,
where $\mathbb{I}(\cdot)$ is the indicator function.  
The final top-$K$ retrieval accuracy for model $M_i$ is then computed by averaging over all stimuli and all other models in the pool: $\mathrm{Acc}_K(i) = \frac{1}{(X-1)N} \sum_{\substack{j=1 \\ j \neq i}}^{X} \sum_{x=1}^{N} \mathrm{Acc}_K^{(i,j)}(x)$
High Top-$K$ accuracy indicates that aligned representations preserve stimulus identity across models. This quantifies the extent to which cross-model representational differences are eliminable once nuisance symmetries are removed. 

\subsection{Datasets}
\label{subsec:app_section_datasets}

\subsubsection{Vision Dataset}

We utilize two primary vision datasets in our experiments: 

\textbf{ImageNet.} ImageNet \cite{deng2009imagenet} is a large-scale image classification dataset comprising over 1.2 million images across 1,000 object categories. It is commonly used to train and evaluate convolutional neural networks and other vision models. For our experiments, we utilized only the ImageNet validation set, which contains 50,000 images. Of these, 45,000 images were used to train the transformation matrices for the pure vision model pools, while the remaining 5,000 images were reserved for inference.

\textbf{THINGS.}The THINGS dataset \cite{hebart2023things} is a curated collection of over 26,000 high-quality, naturalistic object images spanning 1,854 object concepts. Unlike ImageNet, THINGS emphasizes detailed, instance-level representations of everyday objects. We use the THINGS dataset solely for validation. Specifically, we apply the transformation matrices learned from the ImageNet dataset to compute instance-level consistency scores for THINGS. These scores are then compared to human behavioral ratings to assess their correlation with human judgments.

\subsubsection{Language Dataset}
\label{greta-data}

We adopted the English sentence stimulus set introduced by \cite{tuckute2024driving_nathumbehav}. This dataset is well-suited for evaluating universality of language-model representations because it was explicitly curated to span a wide range of linguistic content and (in the original study) a wide range of language-network brain responses, rather than reflecting a single genre or task distribution. All stimuli are short, corpus-extracted six-word sentences. The dataset contains: 

\begin{itemize}
    \item \textbf{Baseline sentences} ($n=1{,}000$): naturalistic six-word sentences curated to maximize linguistic and semantic diversity, combining a semantically diverse subset ($n=534$) with genre-diverse sentences sampled from sources such as news, web media, and transcribed speech ($n=466$).
    \item \textbf{Drive / suppress sentences} ($n=250$ each): out-of-distribution sentences selected (via a GPT2-XL--based encoding model) to elicit maximal (drive) or minimal (suppress) responses in the human language network; identified by ranking candidates from a large-scale search over $\sim$1.8M sentences spanning nine corpora.
\end{itemize}


\paragraph{Sentence-level properties.}
To relate universality scores to interpretable linguistic dimensions, we used the sentence-level properties released alongside the stimulus set (cite~**). These include a model-based predictability measure (surprisal/log probability; estimated with GPT2-XL) and human-rated norms. Specifically, we analyze the following properties:

\begin{itemize}
    \item \textbf{Surprisal / log probability}: model-estimated predictability of the sentence (higher surprisal indicates lower predictability).
    \item \textbf{Grammaticality}: how well the sentence conforms to English grammar.
    \item \textbf{Plausibility}: how much the sentence “makes sense” semantically.
    \item \textbf{Mental-state content}: the extent to which the sentence evokes others' thoughts, beliefs, or intentions.
    \item \textbf{Physical-object content}: the extent to which the sentence evokes physical objects and their interactions.
    \item \textbf{Place content}: the extent to which the sentence evokes places and environments.
    \item \textbf{Valence}: how positive versus negative the sentence content is.
    \item \textbf{Arousal}: how exciting or emotionally activating the sentence content is.
    \item \textbf{Visual imagery}: how easy it is to form a visual mental image of the sentence content.
    \item \textbf{General frequency}: perceived commonness of the sentence in everyday language overall.
    \item \textbf{Conversational frequency}: perceived commonness of the sentence specifically in spoken conversation.
\end{itemize}

\subsubsection{Multimodal Datasets}

\paragraph{MS-COCO.}
MS-COCO is a large-scale image captioning dataset of natural images depicting complex everyday scenes, with each image annotated with five human-authored captions \cite{lin2014microsoft}. We use the official 2017 validation split (Val2017; 5,000 images) and randomly sample 1,000 images with their associated captions for constructing the cross-modal universal embedding space.
 
\paragraph{Flickr8k-Expert.}
Flickr8k-Expert is a human-judgment benchmark for image--caption compatibility, consisting of 1,000 images from the Flickr8k test set paired with 5,822 candidate captions. Each image--caption pair is rated by expert annotators on a 4-point scale (1 = unrelated; 4 = perfect match), with three ratings per pair \cite{kilickaya2017reevaluating,scott2023improved,narins2023vicr}. We use these ratings as ground-truth human judgments to evaluate whether similarity scores derived from the cross-modal universal space track human assessments of caption quality.

\subsubsection{Brain Datasets}

A detailed description of the Natural Scenes Dataset (NSD; http://naturalscenesdataset.org) is provided elsewhere \citep{allen2022massive}. The NSD dataset contains measurements of fMRI responses from 8 participants who each viewed 9,000–10,000 distinct color natural scenes (22,000–30,000 trials) over the course of 30–40 scan sessions. Scanning was conducted at 7T using whole-brain gradient-echo EPI at 1.8-mm resolution and 1.6-s repetition time. Images were taken from the Microsoft Common Objects in Context (COCO) database \citep{lin2014microsoft}, square cropped, and presented at a size of 8.4° x 8.4°. A special set of 1,000 images were shared across subjects; the remaining images were mutually exclusive across subjects. Images were presented for 3 s with 1-s gaps in between images. Subjects fixated centrally and performed a long-term continuous recognition task on the images. The fMRI data were pre-processed by performing one temporal interpolation (to correct for slice time differences) and one spatial interpolation (to correct for head motion). A general linear model was then used to estimate single-trial beta weights. Cortical surface reconstructions were generated using FreeSurfer, and both volume- and surface-based versions of the beta weights were created. In this study, we analyze manually defined regions of interest (ROIs) across both early and higher-level visual cortical areas. For early visual areas, we focus on ROIs delineated based on the results of the population receptive field (pRF) experiment - V1v, V1d, V2v, V2d, V3v, V3d, and hV4. For higher level visual cortex regions, we target the ventral, dorsal, and lateral streams, as defined by the streams atlas.

\subsection{Models}

\subsubsection{Vision Models}

\begin{enumerate}
    \item \textbf{ResNet family} \cite{he2016deep} (ResNet-50, ResNet-34, ResNet-18, ResNet-101):  
    Deep residual networks that use skip connections to mitigate vanishing gradients, enabling very deep convolutional architectures. 

    \item \textbf{ConvNeXt family} (ConvNeXt Small, Tiny, Base, Large) \cite{liu2022convnet}:  
    A modernized CNN design that incorporates architectural ideas from Transformers, such as large kernels and inverted bottlenecks. 

    \item \textbf{Swin Transformers} \cite{liu2021swin} (Swin Small, Tiny, Base, Large):  
    Hierarchical vision Transformers that process images using shifted windows, providing linear computational complexity with respect to image size.

    \item \textbf{Vision Transformers (ViTs)} (ViT Small, Tiny, Base, Large) \cite{dosovitskiy2020image}:  
    Pure Transformer architectures that treat images as sequences of non-overlapping patches, capturing global context through self-attention.

    \item \textbf{DINO models} (DINO Small, Base, ResNet-50) \cite{caron2021emerging}:  
    Self-supervised representations learned via knowledge distillation, producing strong, transferable visual features without labeled data and supporting both ViT and ResNet backbones.

    \item \textbf{VGGs} (VGG 11,13,16,19) \cite{simonyan2014very} - is characterized by its depth, consisting of 16 layers including 13 convolutional layers and 3 fully connected layers. VGG-16 is renowned for its simplicity and effectiveness as well as its ability to achieve strong performance on various computer vision tasks including image classification and object recognition.

    \item \textbf{AlexNet} \cite{krizhevsky2012imagene} - A pioneering CNN that won the 2012 ImageNet competition, introducing ReLU activations, overlapping max-pooling, and dropout for regularization. AlexNet consists of five convolutional layers followed by three fully connected layers
\end{enumerate}

\subsubsection{Language Models}

\begin{enumerate}
    \item \textbf{OpenLLaMA family} (OpenLLaMA-3B, 7B, 13B) \cite{openlm2023openllama,touvron2023llama}:  
    Permissively licensed reproductions of the LLaMA-style decoder-only Transformer architecture, trained as causal language models. We use the 3B/7B/13B checkpoints as representative autoregressive LLMs with matched architecture and varying scale.

    \item \textbf{Qwen2.5 family} (Qwen2.5-0.5B, 1.5B, 3B, 7B, 14B) \cite{qwen2024qwen25}:  
    A modern family of decoder-only Transformer language models spanning a wide range of parameter scales, trained with large-scale pretraining and post-training to support strong general-purpose language understanding and generation.

    \item \textbf{Gemma 2} (Gemma2-2B) \cite{gemma2024gemma2}:
    A lightweight, state-of-the-art decoder-only Transformer model at a practical scale, trained with modern architectural and training refinements.

    \item \textbf{ALBERT family} (ALBERT-base/large/xlarge/xxlarge-v2) \cite{lan2020albert}:  
    Encoder-only Transformer models that introduce parameter-sharing and factorized embeddings to reduce memory while maintaining performance, trained with self-supervised objectives for general-purpose language representations.

    \item \textbf{GPT-2 family} (GPT2-small/medium/large/xl) \cite{radford2019language}:  
    Decoder-only Transformer language models trained with next-token prediction, commonly used as baseline autoregressive LLMs across a range of model sizes.
\end{enumerate}

\subsection{Sanity Checks}
\label{subsec:app_section_1}

\begin{figure*}[h!]
\centering
\includegraphics[width=\linewidth]{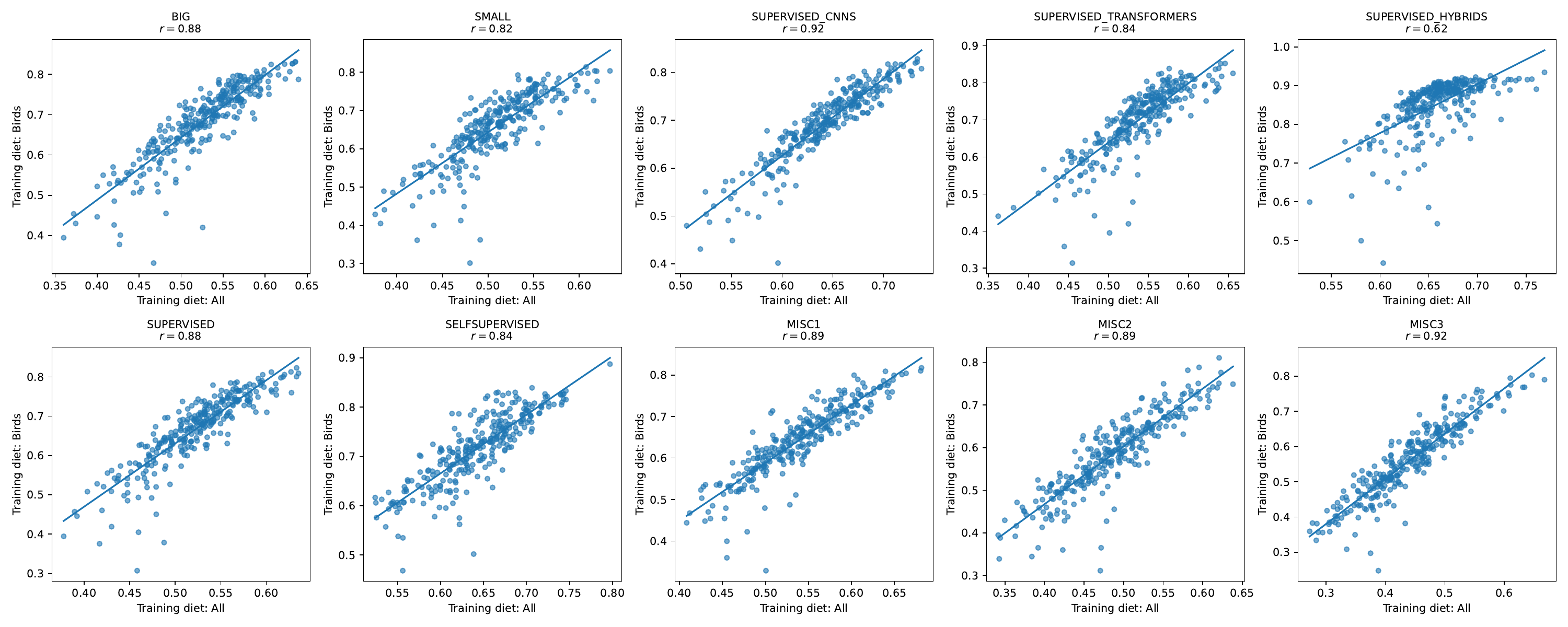}
\caption{Universality similarity scores for test images belonging to the bird parent category from the ImageNet validation set. When the transformation matrices are trained exclusively on bird images, the resulting universality scores are highly correlated with those obtained when transformation matrices are trained on images from all categories.}
\label{fig:app_fig_3}
\end{figure*}

\begin{figure*}[h!]
\centering
\includegraphics[width=\linewidth]{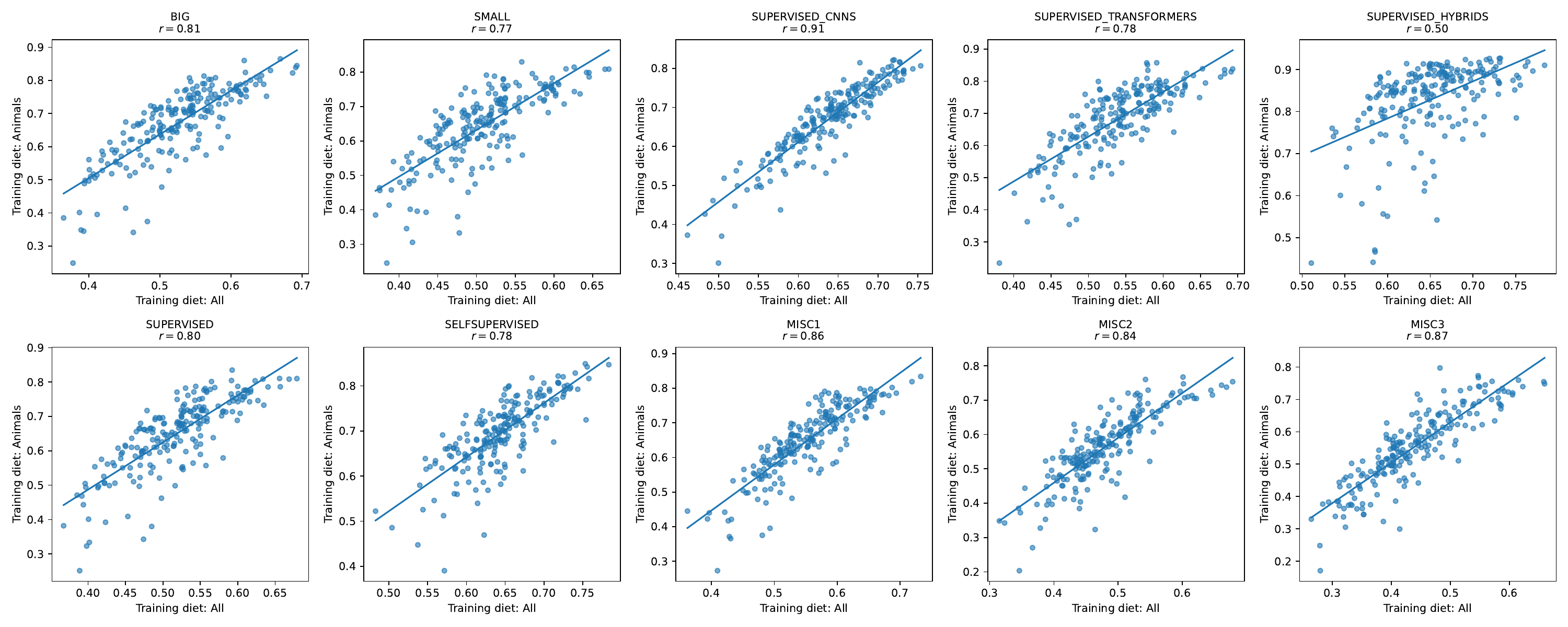}
\caption{Universality similarity scores for test images belonging to the animal parent category from the ImageNet validation set. When the transformation matrices are trained exclusively on animal images, the resulting universality scores are highly correlated with those obtained when transformation matrices are trained on images from all categories.}
\label{fig:app_fig_4}
\end{figure*}

\begin{figure*}[h!]
\centering
\includegraphics[width=\linewidth]{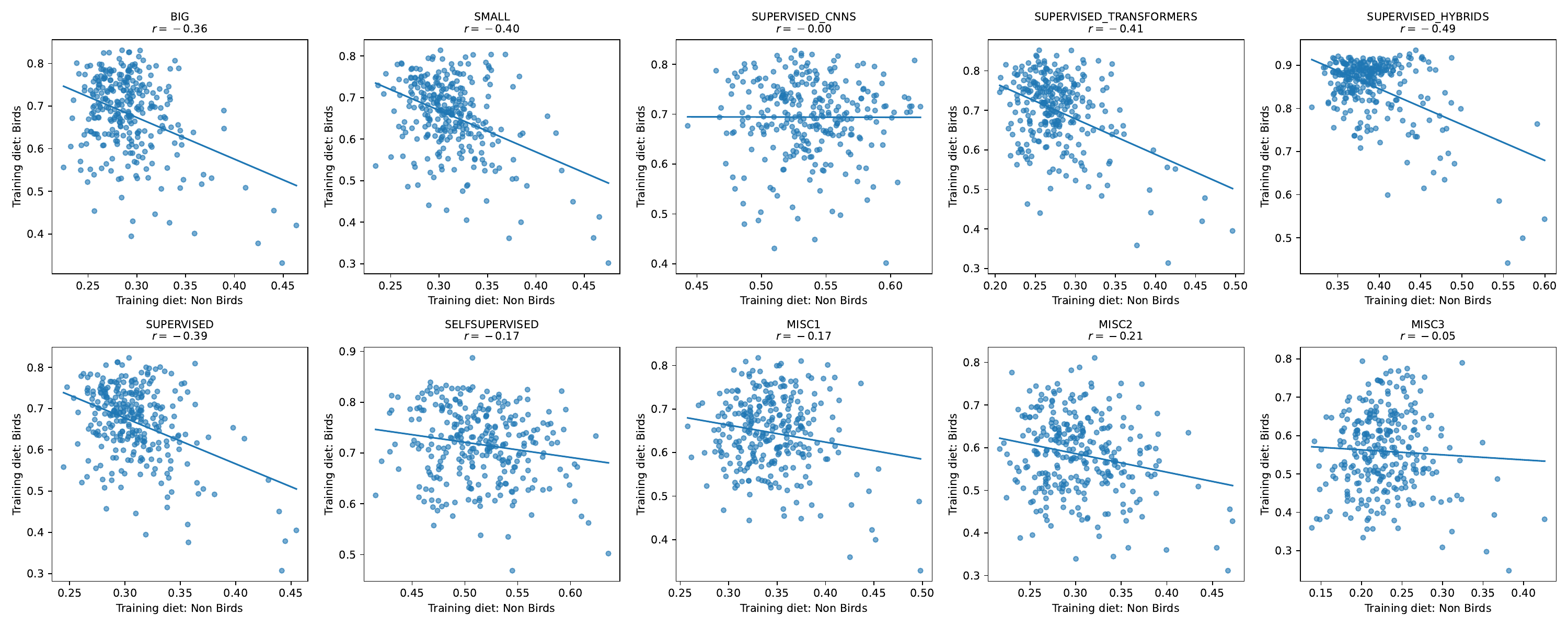}
\caption{Universality similarity scores for test images belonging to the bird parent category from the ImageNet validation set. When the transformation matrices are trained exclusively on bird images, the resulting universality scores are highly uncorrelated with those obtained when transformation matrices are trained on datasets that exclude bird images.}
\label{fig:app_fig_5}
\end{figure*}

\begin{figure*}[h!]
\centering
\includegraphics[width=\linewidth]{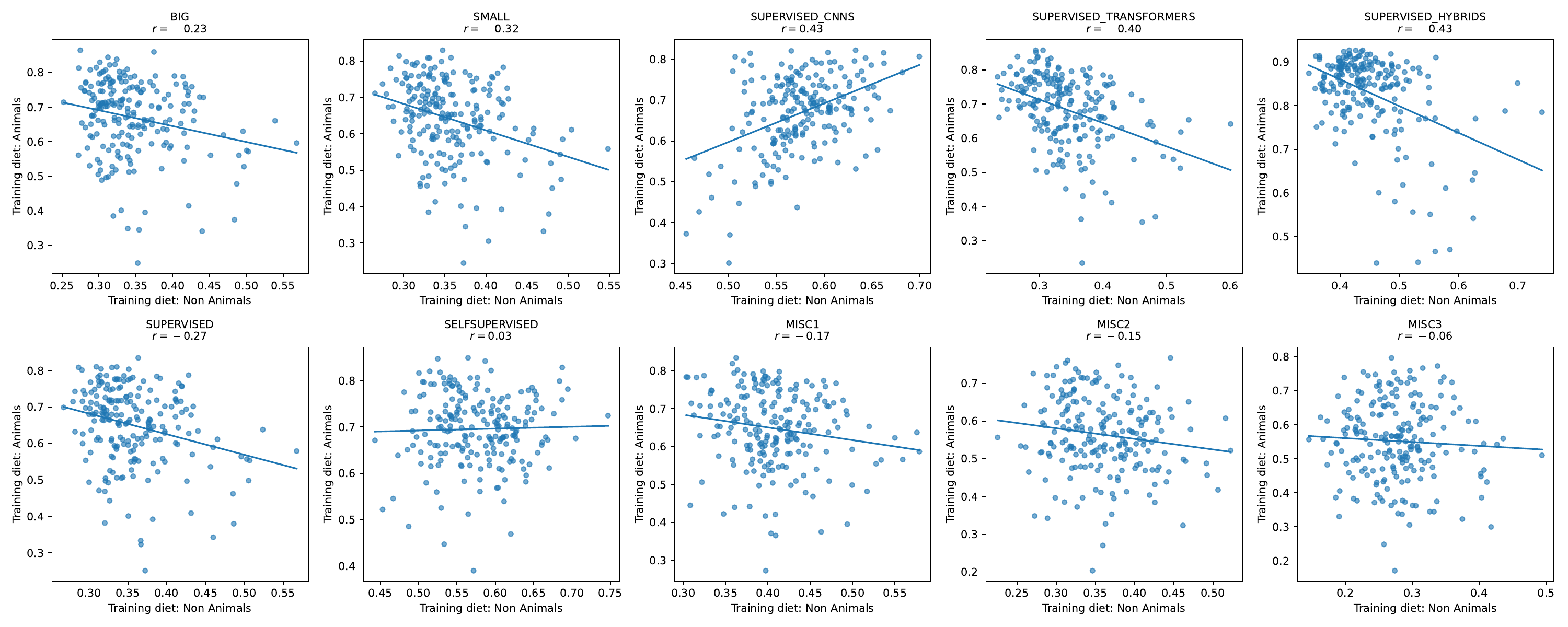}
\caption{Universality similarity scores for test images belonging to the animal parent category from the ImageNet validation set. When the transformation matrices are trained exclusively on animal images, the resulting universality scores are highly uncorrelated with those obtained when transformation matrices are trained on datasets that exclude animal images.}
\label{fig:app_fig_6}
\end{figure*}

We investigate the extent to which the composition of the training stimulus set influences the prediction of universal similarity scores. Specifically, we compare universal similarity scores computed for a given category of test images (e.g., birds, animals) under two training conditions: (i) a category-specific training set containing only images from that category, and (ii) a diverse training set spanning multiple categories, reflecting the standard training regime. We find that universal similarity scores for a given category are highly correlated across these two conditions, indicating that category-level universal similarity is largely invariant to the presence of additional, unrelated categories in the training diet (Figure \ref{fig:app_fig_3}, \ref{fig:app_fig_4}).

We further examine a more stringent setting in which a particular category is entirely absent from the training set. In this case, universal similarity scores for that category are largely uncorrelated with scores obtained when the category is included during training. This result indicates that the method does not generalize reliably to strongly out-of-distribution stimuli (Figure \ref{fig:app_fig_5}, \ref{fig:app_fig_6}).

Taken together, these findings suggest that universal similarity scores are robust to training-set diversity within a category’s support but degrade substantially when evaluated on categories that are completely excluded from the training distribution.


\begin{table}[ht]
\centering
\caption{Average performance across brain model pools (averaged over subjects 1, 2, 5, and 7).}
\label{tab:model_performance_avg}
\footnotesize
\setlength{\tabcolsep}{2.5pt} 

\begin{tabular}{@{}l 
                S[table-format=1.4] 
                S[table-format=1.5] 
                S[table-format=1.4] 
                S[table-format=1.4] 
                S[table-format=1.4] 
                S[table-format=1.4]@{}} 
\toprule
\textbf{Subjects} & {\textbf{Corr. $\uparrow$}} & {\textbf{RMS $\downarrow$}} & {\textbf{Top-1 $\uparrow$}} & {\textbf{Top-5 $\uparrow$}} & {\textbf{Top-10 $\uparrow$}} & {\textbf{Chance-10}} \\
\midrule

\multicolumn{7}{c}{\textit{V1v Model Pool}} \\
\midrule
1 & 0.2976 & 0.0475 & 0.2833 & 0.5133 & 0.6433 & 0.1167 \\
2 & 0.2974 & 0.0473 & 0.3033 & 0.5233 & 0.6267 & 0.1100 \\
5 & 0.2600 & 0.0485 & 0.1800 & 0.3767 & 0.5133 & 0.0967 \\
7 & 0.2515 & 0.0489 & 0.2067 & 0.4167 & 0.5333 & 0.0567 \\

\midrule
\multicolumn{7}{c}{\textit{V1d Model Pool}} \\
\midrule
1 & 0.2835 & 0.0425 & 0.3133 & 0.4833 & 0.5967 & 0.0867 \\
2 & 0.2890 & 0.0419 & 0.3067 & 0.5133 & 0.6233 & 0.1100 \\
5 & 0.2326 & 0.0436 & 0.1433 & 0.3700 & 0.4833 & 0.0833 \\
7 & 0.2326 & 0.0436 & 0.2567 & 0.4267 & 0.5333 & 0.1233 \\
\midrule
\multicolumn{7}{c}{\textit{V2v Model Pool}} \\
\midrule
1 & 0.2449 & 0.0413 & 0.2467 & 0.5200 & 0.6600 & 0.0900 \\
2 & 0.2420 & 0.0408 & 0.2800 & 0.5133 & 0.6400 & 0.1067 \\
5 & 0.2175 & 0.0413 & 0.1367 & 0.3700 & 0.4900 & 0.0833 \\
7 & 0.2291 & 0.0411 & 0.2000 & 0.4633 & 0.5700 & 0.0800 \\
\midrule
\multicolumn{7}{c}{\textit{V2d Model Pool}} \\
\midrule
1 & 0.2344 & 0.0496 & 0.2500 & 0.4433 & 0.5500 & 0.0900 \\
2 & 0.2216 & 0.0499 & 0.1967 & 0.4133 & 0.5600 & 0.1100 \\
5 & 0.1769 & 0.0515 & 0.1167 & 0.2833 & 0.4100 & 0.1233 \\
7 & 0.2024 & 0.0504 & 0.1533 & 0.3567 & 0.5000 & 0.0800 \\
\midrule
\multicolumn{7}{c}{\textit{V3v Model Pool}} \\
\midrule
1 & 0.2318 & 0.0482 & 0.2033 & 0.4767 & 0.6000 & 0.0733 \\
2 & 0.2244 & 0.0482 & 0.2267 & 0.4700 & 0.5833 & 0.1067 \\
5 & 0.1936 & 0.0490 & 0.1133 & 0.3333 & 0.4933 & 0.1167 \\
7 & 0.2013 & 0.0487 & 0.1467 & 0.3667 & 0.4833 & 0.0967 \\

\midrule
\multicolumn{7}{c}{\textit{V3d Model Pool}} \\
\midrule
1 & 0.2338 & 0.0526 & 0.1800 & 0.4400 & 0.5633 & 0.0967 \\
2 & 0.2308 & 0.0527 & 0.2100 & 0.4067 & 0.5567 & 0.0833 \\
5 & 0.1776 & 0.0544 & 0.1100 & 0.3033 & 0.4000 & 0.0800 \\
7 & 0.1981 & 0.0537 & 0.1433 & 0.3267 & 0.4733 & 0.0967 \\
\midrule
\multicolumn{7}{c}{\textit{v4 Model Pool}} \\
\midrule
1 & 0.2047 & 0.0476 & 0.2100 & 0.4500 & 0.5567 & 0.0800 \\
2 & 0.2133 & 0.0467 & 0.2200 & 0.4500 & 0.6067 & 0.0800 \\
5 & 0.1831 & 0.0478 & 0.1733 & 0.3733 & 0.4933 & 0.0867 \\
7 & 0.1801 & 0.0478 & 0.1867 & 0.3933 & 0.5200 & 0.0900 \\

\midrule
\multicolumn{7}{c}{\textit{Ventral Visual Stream Model Pool}} \\
\midrule
1 & 0.1573 & 0.0156 & 0.2433 & 0.5367 & 0.6967 & 0.1000 \\
2 & 0.1668 & 0.0156 & 0.2133 & 0.5300 & 0.6733 & 0.1067 \\
5 & 0.1622 & 0.0155 & 0.1733 & 0.4833 & 0.6467 & 0.1167 \\
7 & 0.1477 & 0.0157 & 0.1833 & 0.4633 & 0.6567 & 0.1067 \\
\midrule
\multicolumn{7}{c}{\textit{Dorsal Visual Stream Model Pool}} \\
\midrule
1 & 0.1540 & 0.0225 & 0.1133 & 0.3100 & 0.4533 & 0.1300 \\
2 & 0.1481 & 0.0225 & 0.1400 & 0.3600 & 0.4900 & 0.0833 \\
5 & 0.1577 & 0.0223 & 0.1200 & 0.3167 & 0.4367 & 0.1133 \\
7 & 0.1390 & 0.0226 & 0.0933 & 0.2833 & 0.4033 & 0.0900 \\
\midrule
\multicolumn{7}{c}{\textit{Lateral Visual Stream Model Pool}} \\
\midrule
1 & 0.1697 & 0.0154 & 0.1467 & 0.4000 & 0.5767 & 0.1033 \\
2 & 0.1782 & 0.0152 & 0.1400 & 0.3800 & 0.5733 & 0.0867 \\
5 & 0.1746 & 0.0153 & 0.1400 & 0.3833 & 0.5600 & 0.0933 \\
7 & 0.1588 & 0.0154 & 0.1300 & 0.3933 & 0.5900 & 0.0833 \\
\bottomrule
\end{tabular}
\end{table}

\begin{figure}[h!]
\centering
\includegraphics[width=\linewidth]{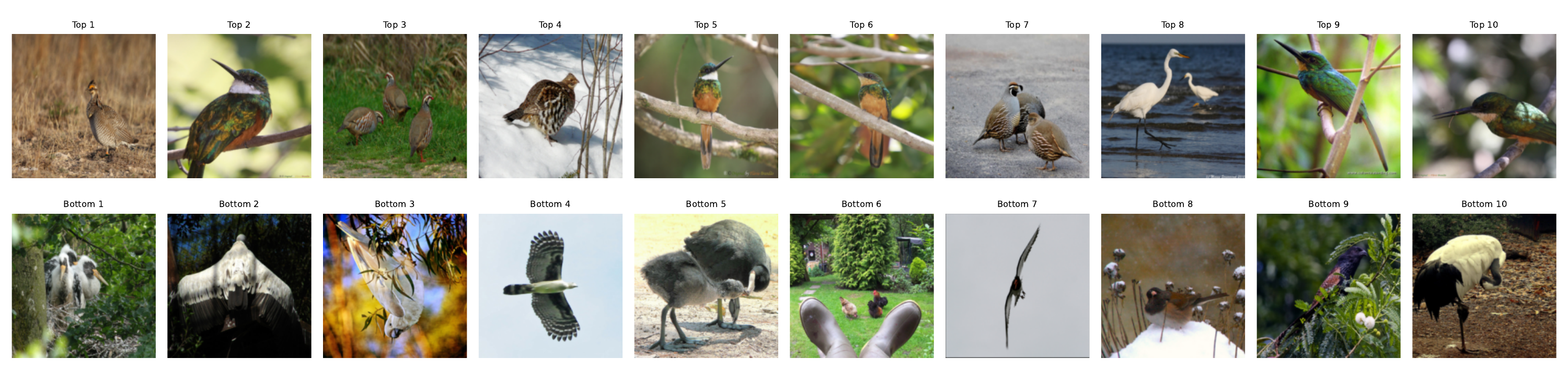}
\caption{Representative images from the Birds parent category showing the highest and lowest universal similarity scores, derived from a pooled ensemble of supervised models with varied architectures and scales.}
\label{fig:app_fig_7}
\end{figure}

\begin{figure}[h!]
\centering
\includegraphics[width=\linewidth]{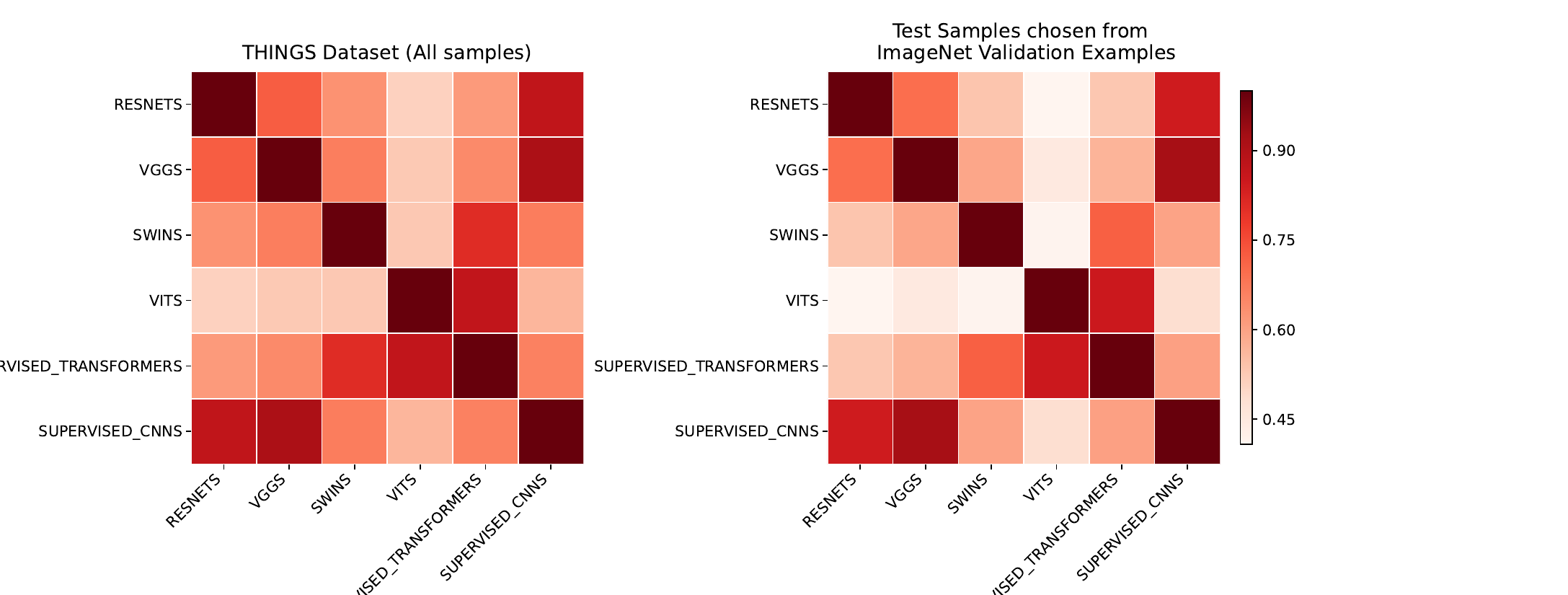}
\caption{Transformation matrices were learned using training samples from the ImageNet validation set to map raw model representations into a universal embedding space. Universal scores were then computed on test samples from ImageNet and on samples from the THINGS dataset using the learned transformation matrices. Consistency across different model pools is shown for THINGS (left) and ImageNet (right). This consistency is highly similar across the two datasets, with a correlation of approximately $0.99$ between the ImageNet- and THINGS-based score similarity matrices, indicating strong generalization to out-of-distribution data.}
\label{fig:app_fig_8}
\end{figure}

\begin{figure}[h!]
\centering
\includegraphics[width=\linewidth]{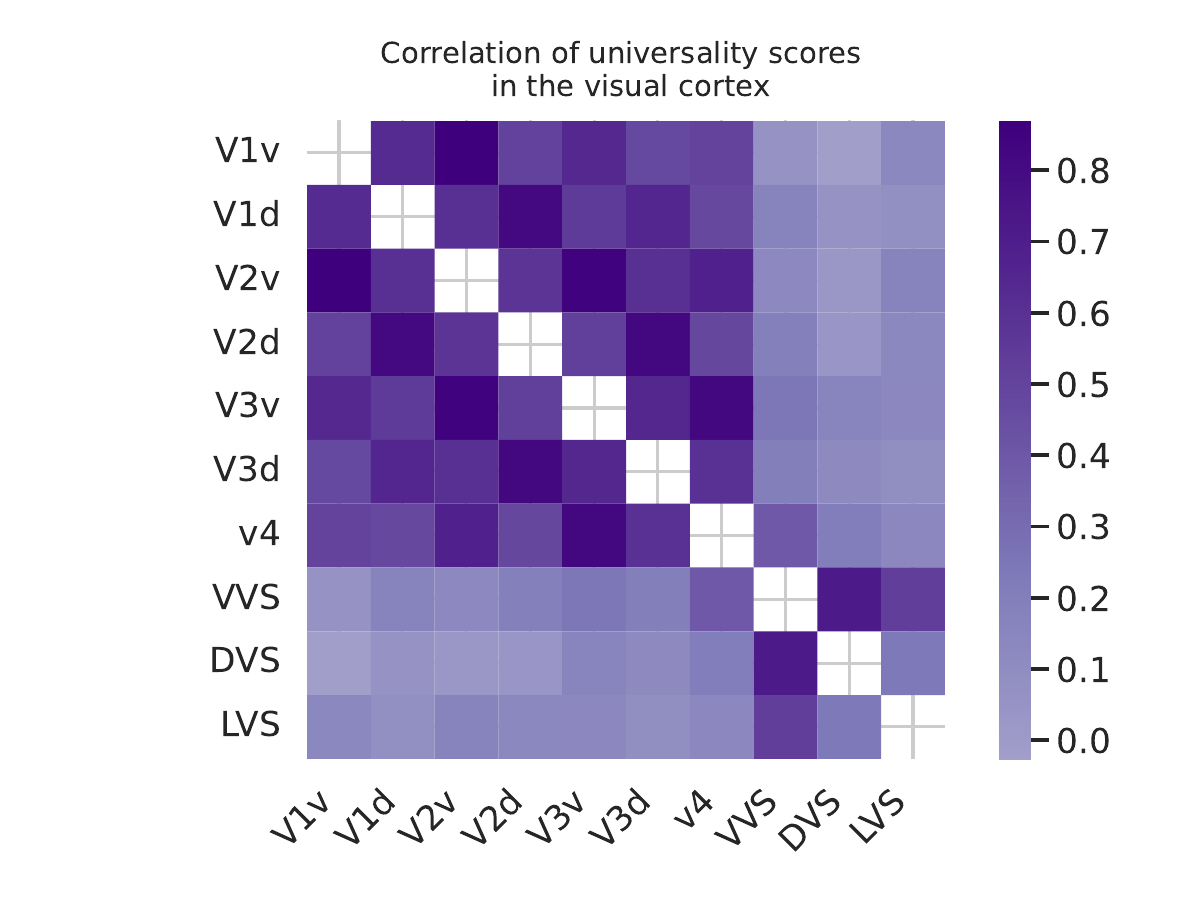}
\caption{Universal similarity scores computed from different biological model pools. Scores within early visual region pools show strong consistency with one another, as do scores within higher-order visual region pools. In contrast, similarity scores between early and higher-order region pools are substantially weaker, reflecting functional differences across the visual hierarchy.}
\label{fig:app_fig_9}
\end{figure}

\begin{figure*}[h!]
\centering
\includegraphics[width=\linewidth]{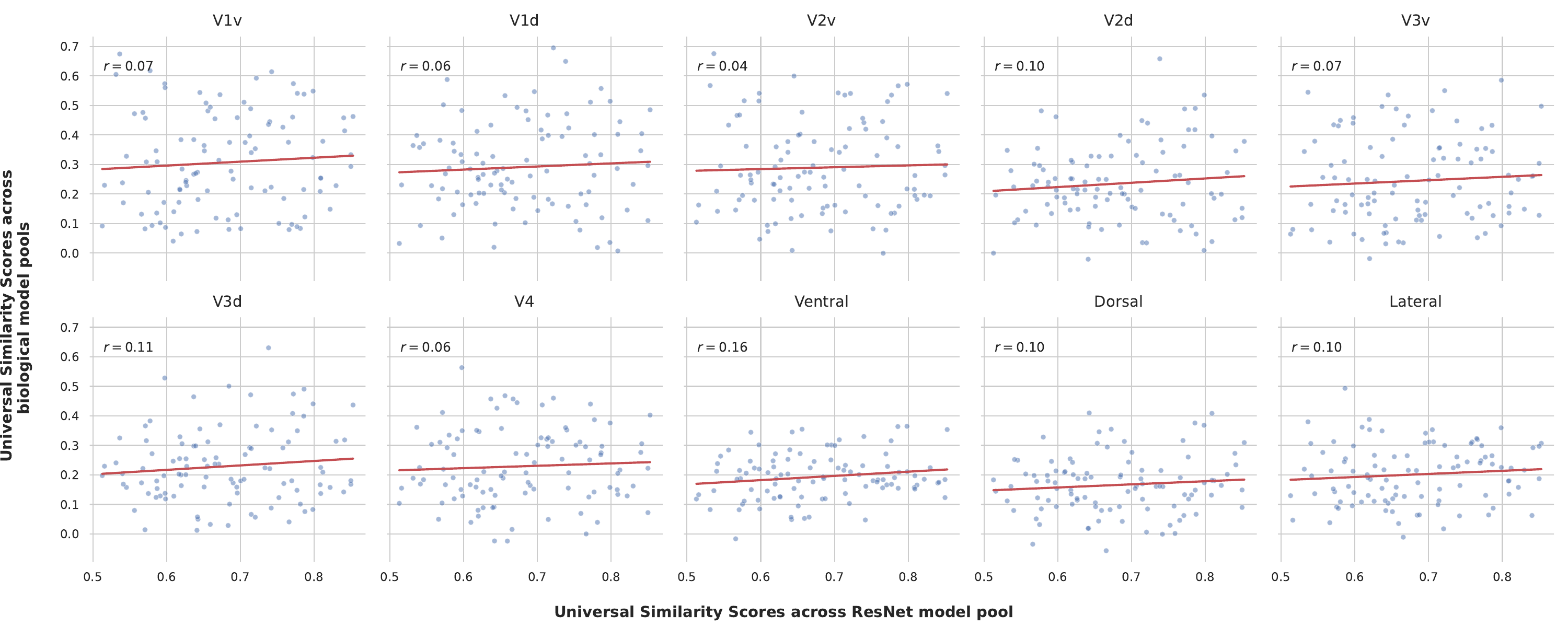}
\caption{Comparing Universal Similarity scores obtained via biological model pools with those obtained via ResNet model pool}
\label{fig:app_fig_10}
\end{figure*}

\begin{table}[t]
\centering
\footnotesize 
\renewcommand{\arraystretch}{1.2}
\setlength{\tabcolsep}{4pt} 
\caption{\textbf{Biological model pools.} Categorized by visual region, architecture, and scale for Procrustes alignment.}
\label{tab:brain-data-groups}

\begin{tabularx}{\columnwidth}{@{} l >{\RaggedRight\arraybackslash}X @{}} 
\toprule
\textbf{Group} & \textbf{Models} \\
\midrule

\rowcolor{tableblue} \multicolumn{2}{l}{\textbf{\textit{Region-based pools}}} \\ 
V1v      & \texttt{V1v - Subjects 1,2,5,7} \\
V1d      & \texttt{V1d - Subjects 1,2,5,7} \\
V2v      & \texttt{V2v - Subjects 1,2,5,7} \\
V2d      & \texttt{V2d - Subjects 1,2,5,7} \\
V3v      & \texttt{V3v - Subjects 1,2,5,7} \\
V3d      & \texttt{V3d - Subjects 1,2,5,7} \\
v4      & \texttt{v4 - Subjects 1,2,5,7} \\
Ventral Visual Stream      & \texttt{Ventral Visual Stream   - Subjects 1,2,5,7} \\
Dorsal Visual Stream      & \texttt{Dorsal Visual Stream   - Subjects 1,2,5,7} \\
Lateral Visual Stream      & \texttt{Lateral Visual Stream   - Subjects 1,2,5,7} \\
\rowcolor{tableblue} \multicolumn{2}{l}{\textbf{\textit{Functionality-based pools}}} \\ 
EARLY      & \texttt{V1v, V1d, V2v, V2d, V3v, V3d, v4 - Subjects 1,2,5,7} \\
LATER      & \texttt{ventral, dorsal, lateral streams - Subjects 1,2,5,7} \\
\bottomrule
\end{tabularx}
\end{table}


\end{document}